\documentclass[10pt,twocolumn,letterpaper]{article}

\usepackage[final]{cvpr}      % To produce the REVIEW version
\usepackage{times}
\usepackage{epsfig}
\usepackage{graphicx}
\usepackage{amsmath}
\usepackage{amssymb}
\usepackage{booktabs}       % professional-quality tables
\usepackage{makecell}
\usepackage{mathtools}
\usepackage{multirow}
\usepackage[accsupp]{axessibility}  % Improves PDF readability for those with disabilities.

\usepackage[pagebackref,breaklinks,colorlinks]{hyperref}

\def\cvprPaperID{2419} % *** Enter the CVPR Paper ID here
\def\confName{CVPR}
\def\confYear{2022}

\usepackage{times,epsfig,graphicx}
\usepackage{algorithm,algorithmic}
\usepackage[dvipsnames]{xcolor}
\usepackage{verbatim}
\usepackage{tabularx}
\usepackage{xspace}
\usepackage{bbm}
\usepackage{bm}

\usepackage{amssymb,amsmath}
\newlength{\smallimage}
        \definecolor{rel}{rgb}{.1,.6,.2}
        \definecolor{nrl}{rgb}{1,1,1}
        \definecolor{qim}{rgb}{1,1,1}

\makeatletter

\def\eg{\emph{e.g.\,}}

\def\ie{\emph{i.e.\,}}

\def\cf{\emph{c.f.\,}}

\def\be{\begin{equation}}
\def\ee{\end{equation}}
\def\bea{\begin{eqnarray}}
\def\eea{\end{eqnarray}}
\def\ben{\begin{eqnarray*}}
\def\een{\end{eqnarray*}}

\def\bi{\begin{itemize}}
\def\ei{\end{itemize}}

\newcommand{\btab}[1]{\begin{tabular}{#1}}
\newcommand{\etab}{\end{tabular}}
\newcommand{\ba}[1]{\begin{array}{#1}}
\newcommand{\ea}{\end{array}}

\DeclareMathOperator*{\argmin}{\mathrm{argmin}}

\def\<{\langle}
\def\>{\rangle}

\newcommand{\R}{\mathbb{R}}

\newcommand{\myparagraph}[1]{\vspace{0.1cm}\noindent\textbf{#1.}}
\newcommand{\myparagraphrw}[1]{\vspace{0.1cm}\noindent\textbf{#1}}



\newcommand{\traincolor}[1]{{\color{NavyBlue}#1}}
\newcommand{\valcolor}[1]{{\color{RubineRed}#1}}
\newcommand{\deploycolor}[1]{{\color{Emerald}#1}}

\usepackage{cuted}          % for the teaser figure

\begin{document}

%%%%%%%%% TITLE
\title{On the Road to Online Adaptation for Semantic Image Segmentation}

\author{
Riccardo Volpi\\
\small{NAVER LABS Europe}\thanks{\href{https://europe.naverlabs.com}{https://europe.naverlabs.com}}
\and
Pau De Jorge\\
\small{NAVER LABS Europe}\\
\small{Oxford University}
\and
Diane Larlus\\
\small{NAVER LABS Europe}
\and
Gabriela Csurka\\
\small{NAVER LABS Europe}
% NAVER LABS Europe\thanks{\href{www.europe.naverlabs.com}{www.europe.naverlabs.com}}\\
% {\tt\small \{name.lastname\}@naverlabs.com}
}

\maketitle

%%%%%%%%% ABSTRACT
\begin{strip}
    
\vspace{-40pt}
\centering
\resizebox{1\linewidth}{!}{%
    \includegraphics{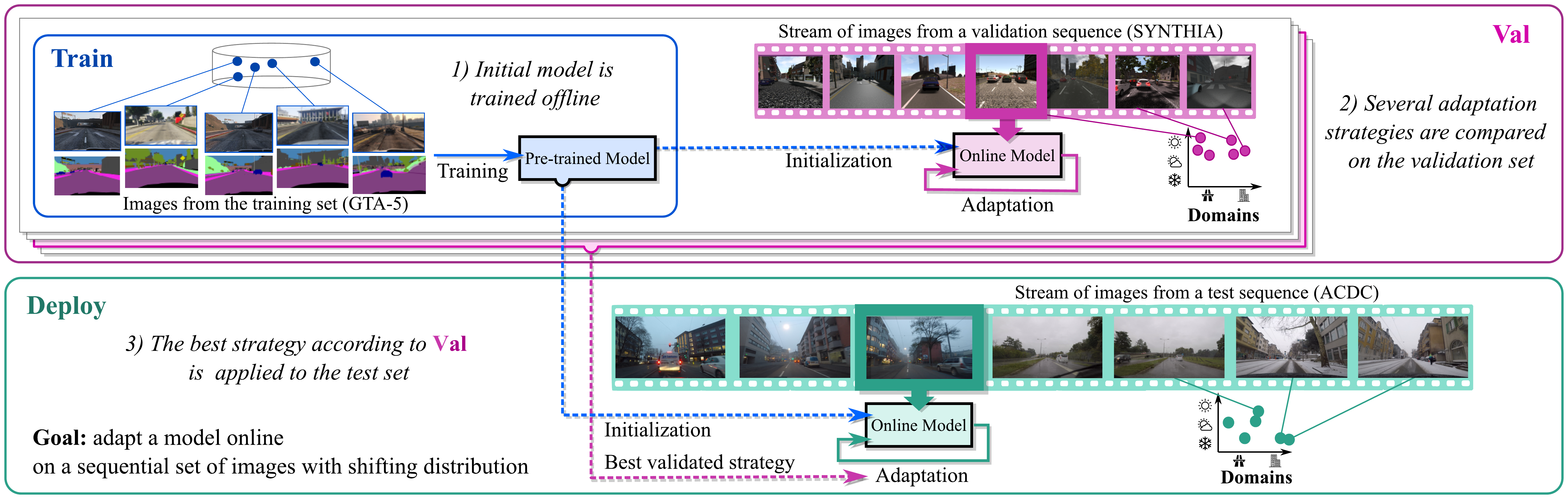}
}
\vspace{-20pt}
\captionof{figure}{
    \textbf{The proposed OASIS benchmark.} We
    formalize a
    domain adaptation 
    task which requires 
    online, unsupervised adaptation 
    of semantic segmentation models
     and propose a novel
    benchmark to tackle it. It is composed of three steps. \traincolor{\textbf{Train:}} A 
    model is trained offline on simulated data (top-left); \valcolor{\textbf{Val:}} Several 
    adaptation strategies
    are validated on simulated data organized in sequentially shifting domains
    (\eg, sunny-to-rainy, highway-to-city), to mimic deploy (top-right).
    \deploycolor{\textbf{Deploy:}} The best 
    validated strategy
    is applied to the test set (bottom).}
\label{fig:front_figure}

\end{strip}

%%%%%%%%% ABSTRACT
\begin{abstract}
We propose a new problem formulation and a corresponding evaluation framework to
advance research on unsupervised domain adaptation for semantic image segmentation.
The overall goal is fostering the development of adaptive learning systems that
will continuously learn, without supervision, in ever-changing environments.
Typical protocols that study adaptation algorithms for segmentation models are
limited to few domains, adaptation happens offline, and human intervention is
generally required, at least to annotate data for hyper-parameter tuning. We argue
that such constraints are incompatible with algorithms that can continuously adapt
to different real-world situations. 
To address this, we propose a protocol where models
need to learn online, from sequences of temporally correlated images, requiring 
continuous, frame-by-frame adaptation. 
We accompany this new protocol
with a variety of baselines to tackle the proposed formulation,
as well as an extensive analysis of their behaviors, 
which can
serve as a starting point for future research. 
\end{abstract}

%%%%%%%%% BODY TEXT

\section{Introduction}\label{sec:intro}

Machine learning systems will often face unfamiliar
conditions when deployed in the real world.
A self-driving car will encounter a new urban environment or unexpected weather; a domestic robot will be deployed in 
a new
house.

In this regard, semantic image segmentation, a
key
task for both examples 
above, is an important case study. Due to the extremely high cost of manually annotating segmentation masks, it is very common to train segmentation models on synthetic data~\cite{RichterECCV16PlayingForData,RosCVPR16SYNTHIADataset};
these models can then be adapted to real environments using domain adaptation techniques~\cite{HoffmanX16FCNsInTheWildPixelLevelAdversarialDA,TsaiCVPR18LearningToAdaptStructuredOutputSpaceSemSegm,HoffmanICML18CyCADACycleConsistentAdversarialDA,ChenCVPR19CrDoCoPixelLevelDomainTransferCrossDomainConsistency,VuCVPR19ADVENTAdversarialEntropyMinimizationDASemSegm,YangCVPR20PhaseConsistentEcologicalDA}.
% Such methods typically adapt a model from 
% a source domain to a target domain: This process happens
Adaptation methods typically run 
offline, require multiple epochs,
and generally
rely on an \textit{annotated validation set}
from the target domain.
Such assumptions are
reasonable for certain scenarios; for instance, consider the problem of adapting a segmentation model for MRI from simulated data
to
real-scanner samples: 
adaptation can happen offline,
and one can ask some experts to provide
at least a few annotated target samples 
for hyper-parameter selection.

Unfortunately, such assumptions do not hold for all scenarios.
In our initial
example of a self-driving car equipped with a semantic
segmentation
module,
once the car is on the road,
samples arrive \textit{sequentially}, \textit{unlabeled}, and
the 
surrounding
environment \textit{continually changes}. 
The segmentation model needs to jointly produce a prediction and continuously adapt as individual images 
arrive.
This is one 
among many such scenarios that require 
unsupervised 
adaptation to happen online, continuously and without the possibility of extensive hyper-parameter tuning on new domains.
We argue that there is a strong need for an 
appropriate
formulation, 
ad hoc
evaluation protocols and
metrics specifically designed for this challenging adaptation problem.

This paper aims at filling this gap.
We introduce a 
constrained problem formulation, which requires \textit{online and
unsupervised adaptation of semantic segmentation models}.
The model is expected to
adapt to new environments by processing 
sequences of temporally correlated frames, from ever changing domains.
Therefore, this formulation does not allow processing large amounts of target samples offline nor performing extensive hyper-parameter search.

Concretely, our \textbf{first contribution} is proposing a three-step \traincolor{train}-\valcolor{val}-\deploycolor{deploy}
evaluation protocol
that emulates as closely as possible the problem of adapting semantic segmentation models as the input distribution shifts.
By design, cross-validating on ``deploy'' samples is not possible.
The three steps are \traincolor{(i)} Pre-training a segmentation model offline on simulated data;
\valcolor{(ii)} Validating the algorithms at hand, \ie carefully choosing all the hyper-parameters -- again on simulated data, yet mimicking the sequential nature of the real world;
\deploycolor{(iii)} Testing the best 
adaptation algorithm
-- according to the previous validation step --  
on realistic, sequential conditions.
Those steps are illustrated in Fig.~\ref{fig:front_figure}.
We formalize this new problem and devise a proper benchmark for \textbf{O}nline \textbf{A}daptation for \textbf{S}emantic \textbf{I}mage \textbf{S}egmentation, \textbf{the OASIS benchmark}.  

Endowed with this framework, our \textbf{second contribution} is tailoring and benchmarking 
a set of adaptation techniques from related fields
for our problem (see Fig.~\ref{fig:rel_work}). 
We also propose different
learning strategies to accommodate the challenges that our protocol brings, 
reporting thorough comparisons among the different methods, and share our main findings, which can serve as a basis for future research on this task. 
As a \textbf{third contribution}, we assess the impact of \textit{catastrophic forgetting} when continuously adapting without supervision and draft a family of methods based on a \textit{reset mechanism} to mitigate 
this issue.
\footnote{Code available at \href{https://github.com/naver/oasis}{https://github.com/naver/oasis}}
\section{Related work}\label{sec:rel_work}

\begin{figure}[t]
	\begin{center}
      \includegraphics[width=0.9\linewidth]{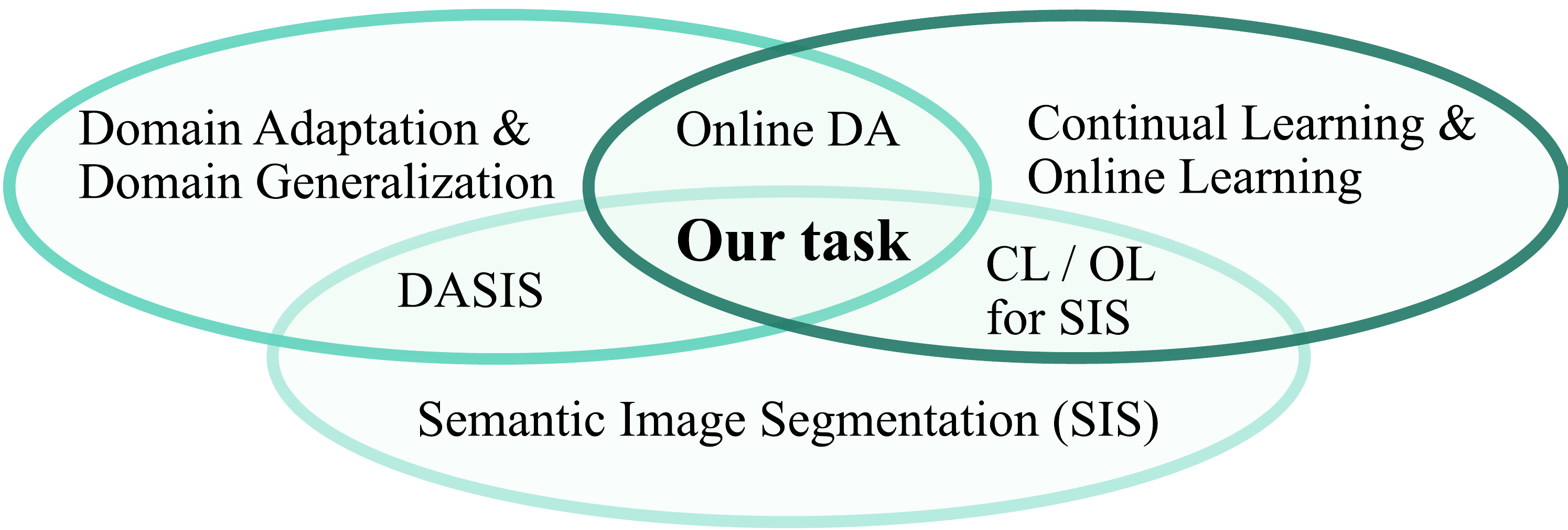}
	\end{center}
	\vspace{-10pt}
	\caption{%\footnotesize 
	\textbf{Our task in the literature.} We focus on the
	task of online unsupervised domain adaptation for semantic image segmentation, which lies at the intersection of several 
	fields.
% 	active topics.
	}
	\vspace{-15pt}
\label{fig:rel_work}
\end{figure}

This paper studies domain adaptation for semantic segmentation
and, hence, relates to both fields. 
Our scenario also has links with the domain generalization and the continual learning literature. Fig.~\ref{fig:rel_work} illustrates 
the connections with different research topics, which we detail below. 

%%%%%%%%% SIS
\myparagraphrw{Semantic image segmentation (SIS)} consists in predicting the class label of every image pixel. 
% Recent 
Modern
methods typically use fully convolutional networks~\cite{LongCVPR15FullyConvolutionalNetworksSegmentation} or recurrent neural networks~\cite{ByeonCVPR15SceneLabelingLSTM,LiangECCV16SemanticObjectParsingGraphLSTM}.  
These models can be further combined with Conditional Random Fields (CRFs)~\cite{ChenICLR15SemanticImgSegmFullyConnectedCRFs,SchwingX15FullyConnectedDeepStructuredNetworks,ZhengICCV15ConditionalRandomFieldsAsRNN,ChandraECCV16FastExactMultiScaleInferenceSemSegmDeepGaussianCRFs}, 
multi-resolution architectures~\cite{LinCVPR17FeaturePyramidNetworksObjDet,GhiasiECCV16LaplacianPyramidReconstructionRefinementSemSegm,ZhaoCVPR17PyramidSceneParsingNetwork,HeCVPR19AdaptivePyramidContextNetworkSemSegm},
or attention mechanisms~\cite{ChenCVPR16AttentionToScaleScalAwarSemSegm,LiBMVC18PyramidAttentionNetworkSemSegm,FuCVPR19DualAttentionNetworkSceneSegm}. 
Encoder-decoder architectures can compress the image into a latent space that captures the underlying semantic information and then decode the latent representation into final predictions~\cite{NohICCV15LearningDeconvolutionNNSegmentation,BadrinarayananPAMI17SegnetDeepConvEncoderDecoder,RonnebergerMICCAI15UNetSegmentation}. 
%
% The most 
A
successful family of SIS methods is DeepLab~\cite{ChenICLR15SemanticImgSegmFullyConnectedCRFs,ChenPAMI17DeeplabSemanticImgSegmentationDeepFullyConnectedCRF,ChenX17RethinkingAtrousConvolutionSemSegm,ChenECCV18EncoderDecoderAtrousSeparableConvSemSegm}, which
combines dilated convolutions for resolution, Atrous Spatial Pyramid Pooling to capture 
%objects  as well as image 
the context at multiple scales, 
and CRFs to refine predictions. 
Lately, it has been shown that transformer architectures are very effective for 
SIS~\cite{ZhengCVPR21RethinkingSiSSeq2SeqPerspTransformers,StrudelICCV21SegmenterTransformerForSemSegm,XieNIPS21SegFormerSemSegmTransformers}.

%%%%%%%% Figure

%%%%%%%%% DA for SIS
\myparagraphrw{Domain adaptation for SIS (DASIS)} 
aims at adapting 
a segmentation model trained on one or several source domains to 
one or more target domains~\cite{ToldoX20UnsupervisedDASSReview,CsurkaX21UnsupervisedDA4SiSComprehensiveSurvey}.
Deep DA~\cite{WangNC18DeepVisualDASurvey,CsurkaX20DeepVisualDomainAdaptation} is particularly relevant for SIS, as the recent approaches are data-hungry and pixel-level annotations are extremely tedious and time consuming to acquire~\cite{CordtsCVPR16CityscapesDataset}.
Since game rendering engines can generate photo-realistic virtual worlds with arbitrary variations in weather, environment and lighting conditions,
synthetic fully-annotated datasets~\cite{RosCVPR16SYNTHIADataset,RichterECCV16PlayingForData}
have become a standard source 
to train SIS models, 
which are then adapted
to real datasets.
Such deep DASIS methods, according to
\cite{ToldoX20UnsupervisedDASSReview,CsurkaX21UnsupervisedDA4SiSComprehensiveSurvey},  can be grouped into adversarial feature  alignment~\cite{HoffmanX16FCNsInTheWildPixelLevelAdversarialDA,HuangECCV18DomainTransferThroughDeepActivationMatching,ZhangCVPR20TransferringRegularizingPredictionSemSegm}, 
output-level adaptation~\cite{TsaiCVPR18LearningToAdaptStructuredOutputSpaceSemSegm,VuCVPR19ADVENTAdversarialEntropyMinimizationDASemSegm,PanCVPR20UnsupervisedIntraDASemSegmSelfSupervision} or image-level style
transfer between domains jointly learned with the segmentation model \cite{HoffmanICML18CyCADACycleConsistentAdversarialDA,MurezCVPR18ImageToImageTranslationDA,SankaranarayananCVPR18LearningSyntheticDataDomainShiftSemSegm,WuECCV18DCANDualChannelWiseAlignmentNetworksUDA,ChangCVPR19AllAboutStructureDASemSeg,YangCVPR20PhaseConsistentEcologicalDA,ChenCVPR19CrDoCoPixelLevelDomainTransferCrossDomainConsistency}.
Self-training and target entropy minimization are also very popular techniques used to improve the adaptation 
\cite{LiCVPR19BidirectionalLearningDASemSegm,ZouECCV18UnsupervisedDASemSegmClassBalancedSelfTraining,ChenICCV17NoMoreDiscriminationCrossCityAdaptationRoadSceneSegmenters,ZouECCV18UnsupervisedDASemSegmClassBalancedSelfTraining,DuICCV19SSFDANSeparatedSemanticFeatureDASemSegm,VuCVPR19ADVENTAdversarialEntropyMinimizationDASemSegm}. 

Differently from these standard 
DASIS methods, we are interested in algorithms that are able to adapt \textit{online}.
These methods cannot be straightforwardly applied in our settings  since we never assume the availability of both source and target empirical distributions.

%%%%%%% DG
\myparagraphrw{Domain generalization (DG)} models learn on one or more source domains with the goal of generalizing to new, unseen domains. 
There is a large diversity of DG methods~\cite{ZhouX20DomainGeneralizationSurvey,WangX20GeneralizingToUnseenDomainsSurveyDG}, 
some of them effectively applied to SIS
\cite{GongCVPR19DLOWDomainFlowAdaptationGeneralization,YueICCV19DomainRandomizationPyramidConsistencySimulationToReal,JinX20StyleNormalizationRestitutionDGDA,VolpiNeurIPS19GeneralizingUnseenDomainsAdvDataAugm,VolpiICCV19AddressingModelVulnerabilityDistrShiftsImageTransf}.
While DG does not take into account model adaptability per-se, 
it is relevant in our context because, 
as we will show, 
starting with stronger representations leads to more effective adaptation results. Hence, DG can play a crucial role for model pre-training to foster adaptability.

\myparagraphrw{Continual learning (CL)} 
consists in gradually enriching 
%gradually enrich
the model as new domains/tasks are encountered while trying 
not to forget about previous ones~\cite{ChenB18LifelongMachineLearning,ParisiNN19ContinualLifelongLearningNNReview}.
In this paper, we assume that the SIS task  -- including the set of semantic labels -- remains unchanged,
%as well as the semantic class labels, 
but the data distribution varies over time
(\eg from \textit{urban environment} to \textit{highway}, or from \textit{sunny} to \textit{rainy}).
Only few methods have addressed this scenario, \eg by 
gradually shifting target domains 
while  replaying old memories to avoid forgetting~\cite{BobuICLRW18AdaptingContinuouslyShiftingDomains}, 
incremental adversarial training and  periodic offline retraining with previously seen samples \cite{WulfmeierICRA18IncrementalAdversarialDAContinualChangingEnvironments}, incrementally adapting the model to changing environments via meta-learning~\cite{VolpiCVPR21ContinualAdaptVisualReprDomainRandMetaLearn} and style transfer \cite{WuICCV19ACEAdaptChangingEnvironmentsSemSegm},
or fine-tuning the domain adapters on the target in a
self-supervised fashion~\cite{PoravITSC19DontWorryAboutWeatherUnsupervisedConditionDependentDA}.
Yet, while these
methods address adaptation to multiple domains
sequentially, none of them considers an online
scenario where the adaptation needs to be performed frame by frame.

\myparagraphrw{Online domain adaptation (ODA)}, 
% as in
related to
online learning (OL)~\cite{CesaBianchiB06PredictionLearningGames}, 
aims at adapting a model when confronted to a stream of data, potentially dealing with one sample at a time % --- and 
(preferably in real time).
% as the model should 
% jointly adapt and predict on-the-fly. 
This is in contrast with incremental DA~\cite{BobuICLRW18AdaptingContinuouslyShiftingDomains,WuICCV19ACEAdaptChangingEnvironmentsSemSegm,VolpiCVPR21ContinualAdaptVisualReprDomainRandMetaLearn}, 
where the model is 
trained to handle 
new domains \textit{offline}.
A few ODA methods have been proposed in the past, in particular to adjust  pre-trained detectors for videos \cite{WangCVPR12DetectionByDetectionsAdaptationForVideo,XuPAMI14DomainDADeformableDPM} or object tracking~\cite{SharmaCVPR12GeneralizedMultiviewAnalysis,GaidonX14SelfLearningCameraObjDetVideo,GaidonBMVC15OnlineDATracking}.
% More recent ODA methods perform online updates of batch normalization (BN)~\cite{IoffeICML15BatchNormAcceleratingDeepNetsTrain} statistics, and are applied to robotic kitting~\cite{ManciniIROS18KittingWildODA} or stereo depth estimation~\cite{Zhang2019OnlineAdaptMetaLearningStereoDepthEstimation}.
Online updates of batch normalization
(BN)~\cite{IoffeICML15BatchNormAcceleratingDeepNetsTrain}
statistics have been shown to help when applied to robotic
kitting~\cite{ManciniIROS18KittingWildODA}.
More recently, online adaptation has been more extensively studied with respect to depth estimation, both in the stereo~\cite{TonioniCVPR19RealTimeSelfAdaptDeepStereo,TonioniCVPR19LearningToAdaptForStereo,ZhangX19OnlineAdaptMetaLearnStereoDepthEst} and in the
monocular~\cite{ZhangCVPR20OnlineDepthLearnAgainstForgettingMonocularVideo,KuznietsovWACV21CoMoDaContinuousMonocularDepthAdaptationUsingPastExp} cases.

% Our formulation requires the model to be adapted both in a \textit{sequential} and \textit{online} manner, and focuses on the challenging SIS task.
Our formulation requires SIS models to be adapted both in a sequential
and online manner.
To the best of our knowledge, there is only one such previous ODA approach for SIS, OnAVOS~\cite{VoigtlaenderBMVC17OnlineAdaptationCNNVideoObjSegm};
yet, 
%this method 
it only considers two classes (moving foreground/background) and assumes the first frame to be fully labeled. In this regard, one of our contributions is to provide a general formulation and companion benchmark to study ODA for SIS
in a realistic framework.

Recently, \textbf{test-time adaptation/training}
techniques have been proposed that can be applied to ODA. Sun et al.~\cite{SunICML20TestTimeTrainingSelfSupervisionGeneralizDistr} optimize 
the model parameters at test time using self-supervised learning.
Wang et al.~\cite{WangICLR21TENTFullyTestTimeAdaptEntropyMin} show that BN
parameters can be optimized by minimizing the entropy of the model's predictions for the target samples. Schneider et al.~\cite{SchneiderNeurIPS2020ImprovingRobustnessCommonCorruptionsCovShiftAdapt} show that mixing BN statistics of pre-trained models and those gathered from the target samples can
significantly boost out-of-domain performance.
In this work, we study different flavors of the latter two approaches, mixing them with different strategies to propose variants 
tailored for our problem formulation.
%\section{Proposed protocol}
\section{The OASIS benchmark}
\label{sec:protocol_oasis}

This paper formalizes the problem of semantic
image segmentation in a constantly changing environment, 
such as the one encountered by self-driving
systems -- 
where adaptation needs to
happen image by image in a sequential fashion. 
With these problems, concrete research questions arise: 
(1) \textit{How well can a model perform without adaptation when exposed to
different conditions?} Tackling this question will help designing  pre-training strategies that adaptation methods should start from.
(2) Assuming that online adaptation at image-by-image level is effective, \textit{should we adapt continuously},
or should the adaptation strategy for each image always start from a pre-trained model? (3) In the case of continual adaptation, \textit{will the model suffer from
catastrophic forgetting}? (4) If it does, \textit{how to overcome it?}

After formalizing the task in Sec.~\ref{sec:problem}, 
in Sec.~\ref{sec:protocol_design} we describe 
\textbf{the OASIS benchmark} with its evaluation protocol, designed to answer the above  questions.

\subsection{Task definition}
\label{sec:problem}

Let $\mathcal{X}^{s} = \{ (x^s_i, y^s_i) \}_{i=1}^{N^s}$ be a 
set of annotated source images, with
$x^s_i \in \R^{H\times W\times 3}$ and $y^s_i \in \R^{H\times W \times C}$. 
Images are of size $(H,W)$
and 
their pixels are
assigned to one of the $C$ semantic classes.
Let  $\mathcal{X}^{t} = ( x^{t}_i )_{i=1}^{N^{t}}$ be a sequence of non-annotated target  samples,  with
$x^{t}_i \in \R^{H\times W\times 3}$.

We define the problem of \textbf{continual unsupervised domain adaptation for semantic image segmentation} starting from the following configuration.  
Let $\mathcal{M}_{\theta_0}$ be a parametric model for SIS that is learned on $\mathcal{X}^{s}$. This model has never seen any target images, not even unlabelled ones. 
Once the adaptation process starts, the model receives unlabelled target samples $x^t_i\sim \mathcal{X}^{t}$,
one at a time.
A given adaptation routine, provided with the current\footnote{One may also consider using a buffer of recent samples or an episodic memory where some previously encountered samples are stored -- but we do not consider this scenario in this work.}
sample $x^t_i$ and the current model $\mathcal{M}_{\theta_{i-1}}$, outputs the adapted model $\mathcal{M}_{\theta_{i}}$.
The latter is used to compute the final predictions $\hat{y}^t_i$.

Given a proper metric -- \eg,
mean-Intersection-over-Union (mIoU)
% , which is typically used to 
% evaluate SIS models 
-- one can compute the performance score on the 
whole sequence $\mathcal{X}^{t}$ by averaging 
single samples' scores, computed by comparing 
$\hat{y}^t_i$ with the ground truth. 
%$y^{t}_i$.
% \footnote{\rict{In practical applications, 
% one may decide to use the model adapted on the previous frame on the current sample 
% $\hat{x}^t_i$, to trade-off accuracy for computational efficiency.}}.

\subsection{Protocol design}
\label{sec:protocol_design}

In order to answer the crucial questions that arise when developing 
% UDA 
adaptive
methods in the real world, we need a carefully designed
evaluation protocol, inherently different from the ones typically %previously 
considered for domain adaptation.

Our core contribution is
{\bf the OASIS benchmark}, 
whose aim is filling this gap.
Its protocol, also depicted in Fig.~\ref{fig:front_figure}, comprises three steps: \traincolor{train}, \valcolor{val}, and \deploycolor{deploy}, which respectively correspond to the \textit{pre-training}, \textit{validation} and \textit{real-world test} components of our benchmark. For those, we build on existing datasets.  
In the following,
we provide the rationale behind each of them and a summary of their implementation (more details in the
%we refer to 
Supplementary
Sec.~\ref{app:dataset}).
%Appendix~\ref{app:dataset}.

\subsubsection{Pre-training \traincolor{(train)}}\label{sec:prot_pretr}

\myparagraph{Rationale} 
Regardless of whether the model should be adapted to new environments or not,
it is crucial for a visual model deployed into the real world to be as robust as possible. The key ingredients to achieve robustness
are generally a strong architecture
and the availability of a large amount of labelled training data.
Domain randomization~\cite{TobinIROS17DomRandTransferDNNSim,YueICCV19DomainRandomizationPyramidConsistencySimulationToReal} and domain generalization~\cite{WangX20GeneralizingToUnseenDomainsSurveyDG,ZhouX20DomainGeneralizationSurvey} 
can be used to boost robustness as well. Our protocol allows to
explore SIS models trained \textit{offline}, on large amounts of
annotated data, and with arbitrary training strategies;
the goal here is providing a strong starting point for further 
deployment.

\myparagraph{Experimental details}
Due to the high cost of human-annotations for SIS, it is common to rely on images rendered via game engines -- as their segmentation annotation comes for free.
Therefore, 
following previous work on DASIS~\cite{HoffmanX16FCNsInTheWildPixelLevelAdversarialDA,TsaiCVPR18LearningToAdaptStructuredOutputSpaceSemSegm}, %
we pre-train our SIS models on GTA-5~\cite{RichterECCV16PlayingForData},
a dataset of $25k$ images that serves as our source domain.

\subsubsection{Validating \valcolor{(val)}}\label{sec:prot_val}

\myparagraph{Rationale}
In order to predict
how the model will behave in real-world environments, it is important to evaluate different adaptation strategies 
in conditions that mimic the real world as closely as possible.
Concretely,
provided with a satisfactory initial model resulting from the previous phase, we need to validate
its out-of-domain performance, as well as the behavior of selected adaptation strategies -- in order to choose the most promising methods and their hyper-parameters.
To this end, we need 
to access or generate a large amount of annotated
\textit{sequences} of
temporally correlated samples\footnote{Sequentiality does not play a role in the
pre-training step, because \textit{offline} training is performed on multiple epochs over the
shuffled dataset.}, with occasional 
sub-domain shifts (weather condition and/or urban environment change).
This validation step outputs the best model and adaptation strategy, together with their tuned  hyper-parameters.

\myparagraph{Experimental details}
We use SYNTHIA~\cite{RosCVPR16SYNTHIADataset} to validate pre-training and adaptation strategies. It comprises simulated sequences from different urban environments (we use \textit{Highway}, \textit{New York-like City}, \textit{European Town}) and different weather/daylight/time-of-the-year conditions (we use \textit{Summer},  \textit{Spring}, \textit{Fall},  \textit{Winter}, \textit{Dawn}, \textit{Sunset}, 
\textit{Night}, \textit{Rain}, \textit{Fog}). 
We design 
episodes where the model needs to process samples from ever changing sequences. We propose a total of $8$ sequences, each built with $5$ random sub-sequences of $300$ frames each, where
both environments and conditions are randomly picked, resulting in several domain shifts. 

\subsubsection{Real-world testing \deploycolor{(deploy)}}\label{sec:prot_test}

\myparagraph{Rationale}
To finally assess performance on realistic situations, the models are tested on real-world sequences
\textit{only once and without further hyper-parameter tuning} -- that is, 
the hyper-parameter setting selected on the validation set must be used for each evaluated method.  

\myparagraph{Experimental details}
We use two datasets built with real images for testing
and comparing the validated models:
Cityscapes~\cite{CordtsCVPR16CityscapesDataset} and
ACDC~\cite{Sakaridis21ACDCDataset}.
Cityscapes comprises
$5k$ real, annotated
images from $50$ cities. We further include sequences with artificial \textit{Fog}~\cite{SakaridisECCV2018ModelAdaptSynthRealDataSemDenseFoggySceneUnderstanding} and \textit{Rain}~\cite{HuCVPR2019DepthAttentionalFeaturesSingleImageRainRemoval}; we will refer to the resulting dataset as Cityscapes A.W. (Artificial Weather), and to the original Cityscapes dataset as Cityscapes O.
ACDC is a recent dataset introduced to study segmentation models on real domain shifts: it contains $4k$
samples from driving sequences uniformly distributed in \textit{Fog}, \textit{Rain},
\textit{Night} and \textit{Snow} conditions. For both datasets we design learning sequences where the model receives samples from a plurality of sequential environments and weather conditions (each contains $4$ different sub-sequences). We have a total of $25$ test sequences, with a number of frames per sequence in the order of hundreds. 
%; see Appendix A for details). %See Supp. Mat. for more detailed information.

\section{Methods}\label{sec:metod}

To tackle the problem we discussed in Sec.~\ref{sec:problem}, we can draw solutions from the literature related to
domain adaptation~\cite{WangNC18DeepVisualDASurvey,CsurkaX20DeepVisualDomainAdaptation,ToldoX20UnsupervisedDASSReview}, the recently
introduced test-time adaptation~\cite{SunICML20TestTimeTrainingSelfSupervisionGeneralizDistr,WangICLR21TENTFullyTestTimeAdaptEntropyMin}, and -- to improve pre-training -- from domain generalization~\cite{ZhouX20DomainGeneralizationSurvey,WangX20GeneralizingToUnseenDomainsSurveyDG}. 
However, finding which of these families of techniques would perform best for our problem of interest is a nontrivial question.

This section details how we tailor a variety of different approaches
from the research fields above %, in order to use them
for the proposed scenario.
When selecting which approaches to start from, we took into account simplicity, computational cost, and memory requirements. While we selected a few representative methods to design our baselines,
some of the components we consider can be replaced by or combined with more complex methods (see Sec.~\ref{sec:rel_work} for some examples).
The same holds for the underlying segmentation model's architecture: while we use
DeepLab-V2~\cite{ChenPAMI17DeeplabSemanticImgSegmentationDeepFullyConnectedCRF}
for all methods (as it is widely used by the DA community~\cite{TsaiCVPR18LearningToAdaptStructuredOutputSpaceSemSegm,VuCVPR19ADVENTAdversarialEntropyMinimizationDASemSegm,LiCVPR19BidirectionalLearningDASemSegm}), any alternative segmentation network could replace it.
Below we briefly describe the chosen baselines, as well as the variants we designed to adapt them to our scenario. %protocol.

\subsection{No adaptation (NA)}
\label{sec:noadapt}

One option to address the proposed scenario is to 
% solely focus 
simply rely
on a 
% single pre-trained 
model pre-trained on the available source data ($\mathcal{M}_{\theta_0}$), 
which can then be used in all circumstances. 
In this case, 
% this model $\mathcal{M}_{\theta_0}$ pre-trained on the available source data 
such model is directly used to predict the segmentation map for each sample $x^{t}_i\sim\mathcal{X}^t$ -- 
without any adaptation strategy. 
In our experiments, 
we consider the two approaches detailed below, which will also be used as 
the starting point for the adaptation
strategies 
defined later on.

\myparagraph{Empirical risk minimization (ERM)} The first model is simply trained via standard ERM, \ie optimizing a loss (typically the pixel-level cross-entropy) using the annotated training set. This is the simplest strategy one can adopt.

\myparagraph{Domain randomization (DR)}
We apply DR~\cite{TobinIROS17DomRandTransferDNNSim}, namely we heavily randomize the appearance of the training samples -- in order to reduce the chances that real-world ones will be perceived as out-of-distribution. While we use simulated data, we do not assume access to the simulator during training; hence, to randomize the training domain, we rely on standard image transformations for data augmentation~\cite{VolpiICCV19AddressingModelVulnerabilityDistrShiftsImageTransf} (find more details in Supplementary Sec.~\ref{app:methods}).
%More details can be found in
%Appendix~\ref{app:methods}.

\subsection{Naive adaptation (N)}
\label{sec:ttl}

In the family of methods presented in this section,
when the model processes a new frame $x^{t}_i$, it always starts the adaptation procedure from the pre-trained source model 
$\mathcal{M}_{\theta_0}$ (ERM or DR). 
We focus on test-time adaptation~\cite{WangICLR21TENTFullyTestTimeAdaptEntropyMin},
self-training via pseudo-labels~\cite{LeeICMLW13PseudoLabelSimpleEfficientSSLMethodDNN} and BN's statistics adaptation~\cite{SchneiderNeurIPS2020ImprovingRobustnessCommonCorruptionsCovShiftAdapt}. We use the suffix \textbf{N-} to indicate approaches from this \textit{Naive} adaptation family.

\myparagraph{N-TENT} This method~\cite{WangICLR21TENTFullyTestTimeAdaptEntropyMin}
adapts the BN's parameters 
(the re-scaling parameters $\beta$ and $\gamma$) by minimizing the average entropy of the target predictions.
Provided with an image $x^t_i$ and a segmentation model $\mathcal{M}_{\theta}$, the method optimizes the following objective:
\vspace{-0.2cm}
\begin{align*}
    \argmin_{\beta,\gamma} \mathcal{L}_H \coloneqq -\sum_{p \in x^t_i} \sum_c^C \hat{y}^p_{i,c} \log \hat{y}^p_{i,c}
\end{align*}
where $\hat{y}^p_{i,c}$ is the prediction for class $c$ of  
pixel $p$ in image $x^t_i$.
Given a fully differentiable model, we can 
optimize this objective via backpropagation~\cite{Rumelhart1986LearnRepresentationsByBackprop}.

\myparagraph{N-PL} Self-training via pseudo-labels (PL)
-- drawn from the semi-supervised
learning literature~\cite{LeeICMLW13PseudoLabelSimpleEfficientSSLMethodDNN} -- is often used in DASIS \cite{LiCVPR19BidirectionalLearningDASemSegm,ZouECCV18UnsupervisedDASemSegmClassBalancedSelfTraining,ChenICCV17NoMoreDiscriminationCrossCityAdaptationRoadSceneSegmenters}.
The idea is to make predictions on the target set (here the current image $x^t_i$), and then use these predictions as ground truth to fine-tune the model.\footnote{While solutions to select only
confident pseudo-labels
have been proposed, 
here we use all pseudo-labels, leaving such studies for the future.}

\myparagraph{N-BN} This method~\cite{SchneiderNeurIPS2020ImprovingRobustnessCommonCorruptionsCovShiftAdapt}
adapts the model's BN statistics by mixing mean and variance stored for each layer 
with the statistics of the current target sample $x^{t}_{i}$:
\begin{align*}
    \mu_l &\coloneqq (1-\alpha)\cdot \mu_l + \alpha \cdot \mathbb{E} \{ F^l(x^{t}_{i})\}\\ 
    \sigma_l^2 &\coloneqq (1-\alpha)\cdot \sigma^2 + \alpha \cdot \mathbb{E} \big\{(F^l(x^{t}_{i})-\mathbb{E}\{F^l(x^{t}_{i})\})^2 \big\} 
\end{align*}
where $ F^l(x^{t}_{i})$ is the feature representation 
of the image $x^{t}_{i}$ at layer $l$ in the network.
Note that typically BN averages across different channels
\textit{and} samples (batch). In our online case, 
as we process a single image at the time, the new mean $\mu_l$ and
variance $\sigma_l^2$ are obtained by averaging only across channels.
The normalization is then carried out as:
\begin{align*}
\widehat{F^l(x^{t}_{i})}=\gamma \cdot \frac{F^l(x^{t}_{i})-\mu_l}{\sigma_l^2} + \beta.
\end{align*}
\myparagraph{N-ST}
Image-to-image (I2I) translation is a popular approach for DASIS
to render target images in a style that resembles the source's or vice-versa~\cite{shrivastava2017learning,WuECCV18DCANDualChannelWiseAlignmentNetworksUDA,ma2018exemplar,cherian2019sem,Richter_2021}. Yet, these methods assume that source and target distributions are known at train time, and an I2I module can be learned offline -- generally, jointly with the segmentation network. On the other hand, 
single-image photorealistic Style Transfer (ST) can be carried out via methods that modify a \textit{content} image to mimic the high-level appearance of a \textit{style} image~\cite{luan2017deep, li2018closed, park2019semantic, yoo2019photorealistic, an2020ultrafast}. In our context, only this latter formulation can be adopted -- assuming the source data is available at deployment.
We test the WCT-2~\cite{yoo2019photorealistic} 
single-image ST algorithm\footnote{We choose this method because it is relatively fast and  performs well, but 
any alternative style transfer method can potentially be adopted.}, by stylizing the current target image with the style of (i) a random source image \textbf{N-ST (random)} or (ii) the closest source image in the feature space \textbf{N-ST (NN)}.

\subsection{Continual adaptation (C)}
\label{sec:contlearning}

In contrast to the methods that we refer to as \textit{Naive}, 
here models are adapted continuously -- image by image -- thus they accumulate knowledge from the target domains over iterations. 
To formalize the difference with their \textit{Naive} counterparts, instead of 
adapting
from $\mathcal{M}_{\theta_0}$ at each stage, \textit{Continual} methods adapt to the frame $x^{t}_i$  starting from the model $\mathcal{M}_{\theta_{i-1}}$ -- which was adapted to the previous frame $x^{t}_{i-1}$.

\myparagraph{Vanilla continual learning}
Some of the methods proposed in the previous section (TENT, PL, BN) can be extended to this scenario, without particular modifications
apart from the specific model the adaptation procedure starts from at each step. We use the suffix \textbf{C-}, to indicate continual learning approaches  (\textbf{C-PL}, \textbf{C-TENT} and \textbf{C-BN}).
Note that \textbf{C-BN}, which continuously 
adapts BN statistics, is  
an established
ODA method~\cite{ManciniIROS18KittingWildODA,ZhangX19OnlineAdaptMetaLearnStereoDepthEst}.

As we will show in Sec.~\ref{sec:experiments},
vanilla continual learning approaches are prone to a severe
vulnerability: if some classes are not encountered
for several frames (\eg, a self-driving car may not encounter pedestrians for a while), continuous
adaptation might
lead to deteriorated performance on such classes
if they are faced again. In extreme cases, the model will
stop predicting some classes altogether.
This is a particular form of \textit{catastrophic forgetting}~\cite{ParisiNN19ContinualLifelongLearningNNReview}, at the intersection between DASIS and continual/online learning. In the following, we propose possible solutions to overcome it.

\myparagraph{Regularizing with source data}
\label{sec:contlearningSA}
We can regularize the chosen adaptation
method, by relying on source data $\mathcal{X}^{s}$
to maintain intact 
the patterns learned during pre-training.
The idea is to ensure that the adapted model maintains the
performance on the source domain -- to avoid forgetting some 
of its categories. 
We achieve this by randomly sampling a small set of 
labeled source samples at each step and adding the cross-entropy
loss w.r.t. these source samples 
to the objective of the adaptation method at hand (\eg, C-PL or C-TENT). This is 
similar to episodic memory approaches in continual
learning~\cite{chaudhryX19tiny}.
We term the corresponding models 
\textbf{C-PL-SR} and \textbf{C-TENT-SR}, respectively.

\myparagraph{Adaptive reset strategies}
We store in memory a copy of the pre-trained model $\mathcal{M}_{\theta_0}$. If at a generic step $i$ the
model $\mathcal{M}_{\theta_i}$ becomes brittle (for example, due to catastrophic forgetting), 
we overwrite the parameters $\mathcal{M}_{\theta_{i}}$ with $\mathcal{M}_{\theta_0}$ before adaptation.
We formalize here a family of methods that \textit{reset} the model parameters whenever 
the model meets some predefined conditions, defined as:
\begin{equation}
\label{eq:reset}
\mathcal{M}_{\theta_i} =
\begin{cases}
  \mathcal{M}_{\theta_0}, & \text{if}\ \psi_{i} > \hat{\psi} \\
  \mathcal{M}_{\theta_{i-1}}, & \text{otherwise}
\end{cases}
\end{equation}
where $\psi_i$ is some metric we track and $\hat{\psi}$ is the corresponding threshold for triggering the reset.

Designing $\psi$ properly is crucial, since we would like 
to only reset if the adapted model is performing worse than the pre-trained one. We propose here two instances of this method -- (i) an oracle and (ii) a workable solution.

\textit{(i) Oracle.} Assuming access to the ground truth, one can reset whenever the adaptive model is performing worse than the pre-trained one.  
Formally, 
given an operator $\mathtt{Perf}$ that evaluates a model's performance (here, the mIoU), we have $\psi_i = \mathtt{Perf}(\mathcal{M}_{\theta_{0}}) - \mathtt{Perf}(\mathcal{M}_{\theta_{i-1}})$ and $\hat{\psi}=0$.
This oracle implementation 
is there to validate
the potential of the reset. We will refer to this strategy as \textbf{Oracle-R-PL} or \textbf{Oracle-R-TENT}, according to the adaptation algorithm used.

\textit{(ii) Discrepancy between predicted classes.} We propose to track the number of classes that $\mathcal{M}_{\theta_0}$ and $\mathcal{M}_{\theta_{i-1}}$ predict over different frames: if the latter consistently predicts a reduced number of classes with respect to the former, we assume that catastrophic forgetting has happened, and we reset the continual model's parameters.
Formally, given an operator $\mathtt{NCl}$ 
that outputs the number of categories predicted in a segmentation map, we have:
\vspace{-0.2cm}
\begin{equation}\label{eq:class}\nonumber
    \psi_{i} = \sum_{j=i-K}^{i}\mathtt{NCl}(\mathcal{M}_{\theta_0}(x^t_j)) - \sum_{j=i-K}^{i}\mathtt{NCl}(\mathcal{M}_{\theta_{i-1}}(x^t_j)),
\end{equation}
where $K$ is the number of previous frames for which we count the number of predicted classes.
We refer to this strategy as \textbf{Class-R-PL} or \textbf{Class-R-TENT}, according to the adaptation algorithm used.
This is a very simple strategy, and more sophisticated reset strategies can be designed.
Our goal here is simply showcasing the effectiveness of this idea 
in its simplest form; developing 
novel reset strategies is part of our future research.
\section{Experiments}\label{sec:experiments}

This section
compares the methods described in Sec.~\ref{sec:metod} on our OASIS benchmark. To recap, \traincolor{(i)}
we pre-train models on GTA-5; \valcolor{(ii)} we validate models and
the  hyper-parameters
of the adaptation strategies on 
learning episodes from SYNTHIA\footnote{A comprehensive description of the different methods' hyper-parameters 
is provided in Supplementary Sec.~\ref{app:methods}.}; \deploycolor{(iii)} we test the best 
methods
on sequences from ACDC, Cityscapes A.W. and Cityscapes O. 

\myparagraph{Experimental details}\label{sec:exp_details}
We use DeepLab-V2~\cite{ChenPAMI17DeeplabSemanticImgSegmentationDeepFullyConnectedCRF} as SIS network, implemented in PyTorch~\cite{PaszkeNeurIPS2019PyTorchPaper}. 
When running TENT- or PL-based adaptation algorithms, we
compute  a single training iteration\footnote{While performing many iterations per frame may be computationally costly for  real systems,
we evaluate this scenario in 
 Supplementary Sec.~\ref{app:results}.}
%Appendix~\ref{app:results}.} 
per image.
%Appendix~\ref{app:methods}.
% together with the code
% for full reproducibility. 
For what concerns the \textbf{evaluation metrics} we rely on, we 
compute the mIoU for each
image encountered in each sequence, averaging over the classes that
are present in each frame -- according to the ground-truth.
At the end of a sequence, we compute the average mIoU across all the images of that sequence.
\begin{table*}[t]
\begin{center}
{\footnotesize
\setlength{\tabcolsep}{3.5pt}
\begin{tabular}{@{}clcccccc@{}}
% \multicolumn {7} {c}{\textbf{Validation (SYNTHIA) and test (ACDC \& Cityscapes) experiments}} \\
\toprule
% \multicolumn{8}{l}{\textit{No adaptation (NA)}} \\
% \midrule
%& & \textbf{Validation} & \multicolumn{3}{c}{\textbf{Test}} \\
& & \textbf{\valcolor{Validation}} & \multicolumn{3}{c}{\textbf{\deploycolor{Test (Deploy)}}} \\
\cmidrule(r){3-3}
\cmidrule(r){4-6}

& \makecell{ \\ \textbf{No adapt. baseline (NA)}} & \makecell{\textbf{SYNTHIA} \\ $39.8$ \tiny{$\pm  3.0$}} 
& \makecell{\textbf{ACDC} \\ $33.6$ \tiny{$\pm 2.5$}} 
& \makecell{\textbf{Cityscapes A.W.} \\ $38.3 $ \tiny{$\pm 2.6$}} 
& \makecell{\textbf{Cityscapes O.}\\ $45.2$ \tiny{$\pm 1.0$}} \\
%\cmidrule(r){2-5}
\midrule
\midrule

% \multirow{2}{*}{\rotatebox[origin=c]{90}{Text}} & row 1 & row 1 \\
%   & row 1 & row 1 \\

& \textbf{Method} & \multicolumn{4}{c}{\textbf{Improvements}} & \textbf{Add. computation} & \textbf{Add. memory}\\
\midrule
\multirow{2}{*}{\rotatebox[origin=c]{90}{\makecell{\textbf{Style}\\\textbf{trans.}}}} 
& N-ST (random) & $+0.7 \%$ \tiny{$\pm 1.7$ } & $-7.4 \%$ \tiny{ $\pm  2.6 $} & $+ 4.1 \%$ \tiny{$ \pm  1.7$ } & $+ 0.4 \%$ \tiny{$ \pm  0.8$ }& ST optim. (++)  & Source set (++)\\
\cmidrule(r){2-8}
& N-ST (NN) & $+0.7 \%$ \tiny{$\pm 1.7$}  & $ -5.1 \% $ \tiny{$ \pm  0.8 $} & $+ 2.9 \%$ \tiny{$ \pm  1.1$ } & $+ 1.0 \%$ \tiny{ $\pm  0.3$ }& ST optim. \& NN  (+++)  & Source set (++) \\

% \textit{\textbf{Naive adaptation}} & & &  &&\\
\midrule
% \midrule
\multirow{3}{*}{\rotatebox[origin=c]{90}{\makecell{\textbf{Naive}\\\textbf{adapt.}}}}
& N-BN  & $+2.7 \%$ \tiny{$\pm 0.8$ } & $+ 2.4 \%$ \tiny{$ \pm  0.6$ } & $+ 1.9 \%$ \tiny{$ \pm  0.8$ } & $+ 1.2 \%$ \tiny{$ \pm  0.1$ } & BN stat. update (*) & - \\
\cmidrule(r){2-8}
&N-PL & $+3.5 \%$ \tiny{$\pm 1.0$ }  & $+ 2.9 \% $\tiny{$ \pm  0.6$ } & $+ 2.4 \%$ \tiny{$ \pm  1.0$ } & $\mathbf{+ 1.4 \%}$ \tiny{ $\mathbf{\pm  0.2} $} & $\mathcal{O}(\mathtt{train steps})$ (+) & - \\
\cmidrule(r){2-8}
&N-TENT & $\mathbf{+8.5 \%}$ \tiny{$\mathbf{\pm 3.1}$ } & $+ 4.9 \%$ \tiny{ $\pm  2.0$ } & $+ 3.1 \%$ \tiny{$ \pm  3.6$ } & $-1.2 \%$ \tiny{$ \pm  0.7$ } & $\mathcal{O}(\mathtt{train steps})$ (+) & - \\
\midrule
% \midrule
% \textit{\textbf{CL adaptation}} & & &  &&\\
\multirow{3}{*}{\rotatebox[origin=c]{90}{\makecell{\textbf{CL}\\\textbf{Vanilla}}}}
&C-BN  & $+6.1 \%$ \tiny{$\pm 3.7$ }  & $+ 6.8 \% $ \tiny{ $\pm  3.6$ } & $+ 7.7 \% $ \tiny{$ \pm  4.3 $ } & $-0.1 \%$ \tiny{$ \pm  1.4$ } & BN stat. update (*) & - \\
\cmidrule(r){2-8}
&C-PL & $-19.9 \%$ \tiny{$\pm 12.0$ } & $-11.7 \% $ \tiny{$ \pm  8.1$ } & $-9.4 \%$ \tiny{$ \pm  8.9 $} & $-17.4 \%$ \tiny{$ \pm  3.2 $} & $\mathcal{O}(\mathtt{train steps})$ (+) & - \\
\cmidrule(r){2-8}
&C-TENT & $+2.5 \%$ \tiny{$\pm 6.8$ }  & $+ 2.7 \%$ \tiny{$ \pm  6.7 $} & $+ 6.4 \%$ \tiny{$ \pm  5.6 $} & $-0.9 \%$ \tiny{ $\pm  1.2$ } & $\mathcal{O}(\mathtt{train steps})$ (+) & - \\
\midrule
% \midrule
% \textit{\textbf{CL+SR adaptation}} & & &  &&\\

\multirow{2}{*}{\rotatebox[origin=c]{90}{\makecell{\textbf{CL}\\\textbf{SrcReg}}}}
&C-PL-SR & $+4.9 \%$ \tiny{$\pm 3.9$ }  & $+ 2.8 \% $\tiny{ $\pm  2.9$ } & $+ 3.9 \%$ \tiny{$ \pm  3.2 $} & $+ 0.5 \%$ \tiny{$ \pm  0.5$ } & $\mathcal{O}(\mathtt{train steps})$ (+)& Source set (++)\\
\cmidrule(r){2-8}
&C-TENT-SR  & $+7.2 \%$ \tiny{$\pm 4.0$ } & $+ 5.8 \%$ \tiny{$ \pm  3.7 $} & $+ 4.7 \%$ \tiny{$ \pm  3.7$ } & $+ 0.2 \%$ \tiny{ $\pm  0.5 $}& $\mathcal{O}(\mathtt{train steps})$ (+)& Source set (++)\\
\midrule
% \midrule
% \textit{\textbf{Class-reset adaptation}} & & & & & \\

\multirow{2}{*}{\rotatebox[origin=c]{90}{\makecell{\textbf{CL}\\\textbf{Reset}}}}
&Class-R-PL  & $+7.2 \%$ \tiny{$\pm 3.9$ } & $\mathbf{+ 8.2 \%}$ \tiny{$\mathbf{ \pm  3.4 }$} & $\mathbf{+ 9.0 \%}$ \tiny{$\mathbf{ \pm  5.1 }$} & $+0.0 \%$ \tiny{$ \pm  1.4$ }& $\mathcal{O}(\mathtt{train steps})$  (+)& Backup net (+)\\
% \midrule
% Dist-N-PL  & $+6.3 \%$ \tiny{$\pm 4.3$ } & $+ 4.9 \%$ \tiny{$ \pm  4.0 $} & $+ 7.4 \%$ \tiny{$ \pm  5.0 $} & $+ 0.1 \%$ \tiny{$ \pm  1.4 $}& $\mathcal{O}(\mathtt{train steps})$ & Backup net\\
\cmidrule(r){2-8}
&Class-R-TENT & $\mathbf{+8.3 \%}$ \tiny{$\mathbf{\pm 4.2}$ }  & $+ 7.3 \% $ \tiny {$ \pm  3.9 $ } & $\mathbf{+ 9.1 \%} $ \tiny {$\mathbf{ \pm  4.9 }$ } & $+ 0.9 \% $ \tiny {$ \pm  1.3 $ } & $\mathcal{O}(\mathtt{train steps})$ (+) & Backup net (+)\\
\midrule
\midrule
% \textit{\textbf{Oracle-reset adaptation}} & & &  &&\\

\multirow{2}{*}{\rotatebox[origin=c]{90}{\makecell{\textbf{CL}\\\textbf{Oracle}}}}
&Oracle-R-PL  & $+10.8 \%$ \tiny{$\pm 4.5$ }  & $+ 11.6 \% $ \tiny{$ \pm  3.8$ } & $+ 12.7 \%$ \tiny{$ \pm  5.6$ } & $+ 2.9 \%$ \tiny{$ \pm  1.4 $}& $\mathcal{O}(\mathtt{train steps})$ (+) & Backup net (+)\\
\cmidrule(r){2-8}
&Oracle-R-TENT  & $+11.4 \%$ \tiny{$\pm 4.4$ }  & $+ 10.9 \%$ \tiny{ $\pm  4.1$ } & $+ 12.2 \%$ \tiny{$ \pm  5.9$ } & $+ 1.9 \%$ \tiny{ $\pm  1.4 $}& $\mathcal{O}(\mathtt{train steps})$ (+) & Backup net (+)\\
\bottomrule
\end{tabular}
}
\end{center}
\vspace{-12pt}
\caption{\footnotesize Results associated with SYNTHIA~\cite{RosCVPR16SYNTHIADataset} %sequences, used to cross-validated hyper-parameters (column $2$), and with 
ACDC~\cite{Sakaridis21ACDCDataset}, Cityscapes  Original (O.)~\cite{CordtsCVPR16CityscapesDataset} and Cityscapes Artificial Weather (A.W.) \cite{SakaridisECCV2018ModelAdaptSynthRealDataSemDenseFoggySceneUnderstanding,HuCVPR2019DepthAttentionalFeaturesSingleImageRainRemoval}  sequences. %(columns $3-5$). 
Hyper-parameters used for ACDC, Cityscapes O. and Cityscapes A.W. are the ones selected via SYNTHIA experiments. Results are reported as average percentage improvements over the non-adapted baselines obtained with the pre-trained model $\mathcal{M}_{\theta_0}$
(see NA results, on top). For all experiments, we use DR$\uparrow\uparrow$ as pre-trained model. The final two columns provide insights about
%report information related to 
additional computations and additional memory requirements over the NA baseline;
the symbols (+)/(++)/(+++) provide qualitative orderings for the different methods (increasing overhead).
%, respectively.
*Updating BN statistics requires neglectable computations. 
Bold numbers indicate best results for each dataset; different numbers per column are highlighted in case of comparable results.
}
\label{tab:all_multi}
\vspace{-10pt}
\end{table*}
\begin{table}[t]
\begin{center}
{\footnotesize
\setlength{\tabcolsep}{3.5pt}
\begin{tabular}{@{}lcccc@{}}
\multicolumn {5} {c}{\textbf{Effect of \traincolor{pre-training} w/ domain randomization (DR) on NA}} \\
\toprule
% & \multicolumn{3}{c}{\textbf{Benchmark}} \\
% \cmidrule(r){2-4}
\textbf{Training} & \textbf{SYNTHIA} & \textbf{ACDC} & \textbf{Cityscapes A.W.} & \textbf{Cityscapes O.} \\
\midrule
ERM & $35.9 \pm 2.5 $ & $ 29.5 \pm 2.5 $ & $ 35.6 \pm 1.9 $ & $ 40.3 \pm 0.9$ \\
\midrule
DR$\uparrow$ & $34.3 \pm 3.3 $ & $ 29.5 \pm 2.4 $ & $ 36.2 \pm 2.3 $ & $ 41.2 \pm 1.0 $\\
\midrule
DR$\uparrow\uparrow$ & $\mathbf{39.8 \pm 3.0} $ & $ \mathbf{33.6 \pm 2.5} $ & $ \mathbf{38.3 \pm 2.6} $ & $ \mathbf{45.2 \pm 1.0}$\\
\midrule
DR$\uparrow\uparrow\uparrow$ & $31.9 \pm 3.0 $ & $ 26.7 \pm 2.3 $ & $ 33.2 \pm 2.5 $ & $ 37.7 \pm 1.1$ \\
% \midrule
% \midrule
% ERM from~\cite{AdaptSegNetCodebase} & & & \\
\bottomrule
\end{tabular}
}
\end{center}
\vspace{-12pt}
\caption{\footnotesize 
No adaptation (NA) results obtained by the pre-trained model $\mathcal{M}_{\theta_0}$, trained via standard ERM or DR, and applied to %learning episodes from 
SYNTHIA~\cite{RosCVPR16SYNTHIADataset},
ACDC~\cite{Sakaridis21ACDCDataset}, and
Cityscapes~\cite{CordtsCVPR16CityscapesDataset} (Artificial Weather and Original) % datasets 
-- in mIoU. 
% We also report performance obtained by using an ERM model downloaded from a third-party repo~\cite{AdaptSegNetCodebase}, for reference. 
%The up arrows 
The $\uparrow$ indicate the severity of randomization  (image transformations) applied in each training batch
(see Supplementary Sec.~\ref{app:methods}.).
%Appendix~\ref{app:methods}). 
Note that we compute mIoU results per image, %following our protocol's sequences, 
and compute the average at the end (our own protocol): these results should not be compared with those reported in standard DASIS~\cite{ToldoX20UnsupervisedDASSReview}.}
\label{tab:dom_rand}
\vspace{-10pt}
\end{table}

\myparagraph{Main results and takeaways}\label{sec:main_res}
We report in Table~\ref{tab:all_multi} the average
results for each dataset, together with details about
% additional 
computational and memory overheads with
respect to the non-adapted model (NA baseline). More
concretely, we 
report the average gain (in \%) over the NA baseline and
the standard deviation computed over the different
sequences. All results  were obtained with the
hyper-parameters selected on SYNTHIA, without any 
hyper-parameter tuning on 
ACDC, Cityscapes A.W and Cityscapes O.  

As a first general observation, we see that several methods carry statistically significant gains over the NA baseline for SYNTHIA, ACDC and Cityscapes A.W.
Importantly, we can appreciate that methods that rank best on SYNTHIA, in general also rank best on the test sets, showing that the validation sequences
of
our benchmark allow validating models and algorithms that transfer to real-world sequences. This is a  crucial point, because in practice we cannot assume to be able to cross-validate the methods for
every new environment the
models are deployed in.

A peculiar case is Cityscapes O.:
% , where
here, excluding the Oracle-R approach, none of the methods brings significant improvements. This can be due to the fact that i) the baseline model (DR $\uparrow\uparrow$) is already strong (\cf performance across the different benchmarks in Table~\ref{tab:dom_rand}) and ii) there is not much weather, lighting or even environment changes in this dataset despite the fact that images have been acquired in different cities (see examples in
Supplementary Sec.~\ref{app:dataset}).
%Appendix~\ref{app:dataset}).

The rest of the section analyzes the methods we have evaluated. We provide our \textit{main observations} below:
%\vspace{0.1cm}

\myparagraph{Domain randomization improves the robustness of the initial model}
Table~\ref{tab:dom_rand} compares the performance of SIS models trained  without data augmentation (ERM) \vs the model trained with increasing level
of  domain randomization (DR). 
We see that, if the randomization level
is  properly selected on the validation set (SYNTHIA), the corresponding best DR model (third row)
not only significantly outperforms ERM (first row) on all test datasets, but also the  models trained with the  other randomization levels, showing good generalization.
Therefore, for all adaptation experiments (Table~\ref{tab:all_multi}) we use $\mathcal{M}_{\theta_0}$ corresponding to  DR$\uparrow\uparrow$.

It is of utmost importance to start from a strong initial source model, \ie
we should use the best-performing NA baseline. For instance, applying N-TENT to an ERM model on ACDC leads to an average mIoU performance of $31.5$, lower than the one achieved with the NA(DR$\uparrow\uparrow$) model. 
Using weak baselines can lead to misleading results,
as also shown by Gulrajani and Lopez-Paz~\cite{GulrajaniICLR21InSearchOfLostDomGen} in the context of domain generalization.

\myparagraph{Single-image style transfer does not seem sufficient for adaptation}
While in general style transfer 
helps DASIS both as a pre-processing~\cite{LiIJCAI17DemystifyingNeuralStyleTransfer,CsurkaTASKCV17DiscrepancyBasedNetworksUnsupervisedDAComparativeStudy,ThomasACCV19ArtisticObjectRecognitionUnsupervisedStyleAdaptation} or learned 
jointly with the segmentation on the whole dataset~\cite{HoffmanICML18CyCADACycleConsistentAdversarialDA,MurezCVPR18ImageToImageTranslationDA,ChenCVPR19CrDoCoPixelLevelDomainTransferCrossDomainConsistency,MustoBMVC20SemanticallyAdaptiveI2ITranslationDASemSegm}, single-image style transfer from source to target 
% in our case 
leads to marginal improvements over the baselines in most of 
our experiments,
and significantly degrades performance on ACDC. 
For what concerns the latter, we observed that ST can still be effective when adapting to \textit{Night} sequences, probably due to the fact that
the appearance change is drastic (Supplementary
Sec.~\ref{app:results}). 

\myparagraph{Batch norm's statistics adaptation generally helps}
It is known for DA~\cite{ManciniIROS18KittingWildODA} and for model robustness~\cite{SchneiderNeurIPS2020ImprovingRobustnessCommonCorruptionsCovShiftAdapt} that
adapting the BN statistics is a simple yet effective 
way to improve model performance. This is also confirmed 
in our case, as shown in Table~\ref{tab:all_multi}, where 
BN adaptation always brings improvements over the NA baseline -- 
both when applied independently frame-by-frame (N-BN) or in a cumulative manner when adapting continuously (C-BN). 
In Supplementary Sec.~\ref{app:results},
we provide further analyses on the impact of mixing BN statistics at test-time~\cite{SchneiderNeurIPS2020ImprovingRobustnessCommonCorruptionsCovShiftAdapt}. 

\begin{figure}
	\begin{center}
     \includegraphics[width=1.0\linewidth]{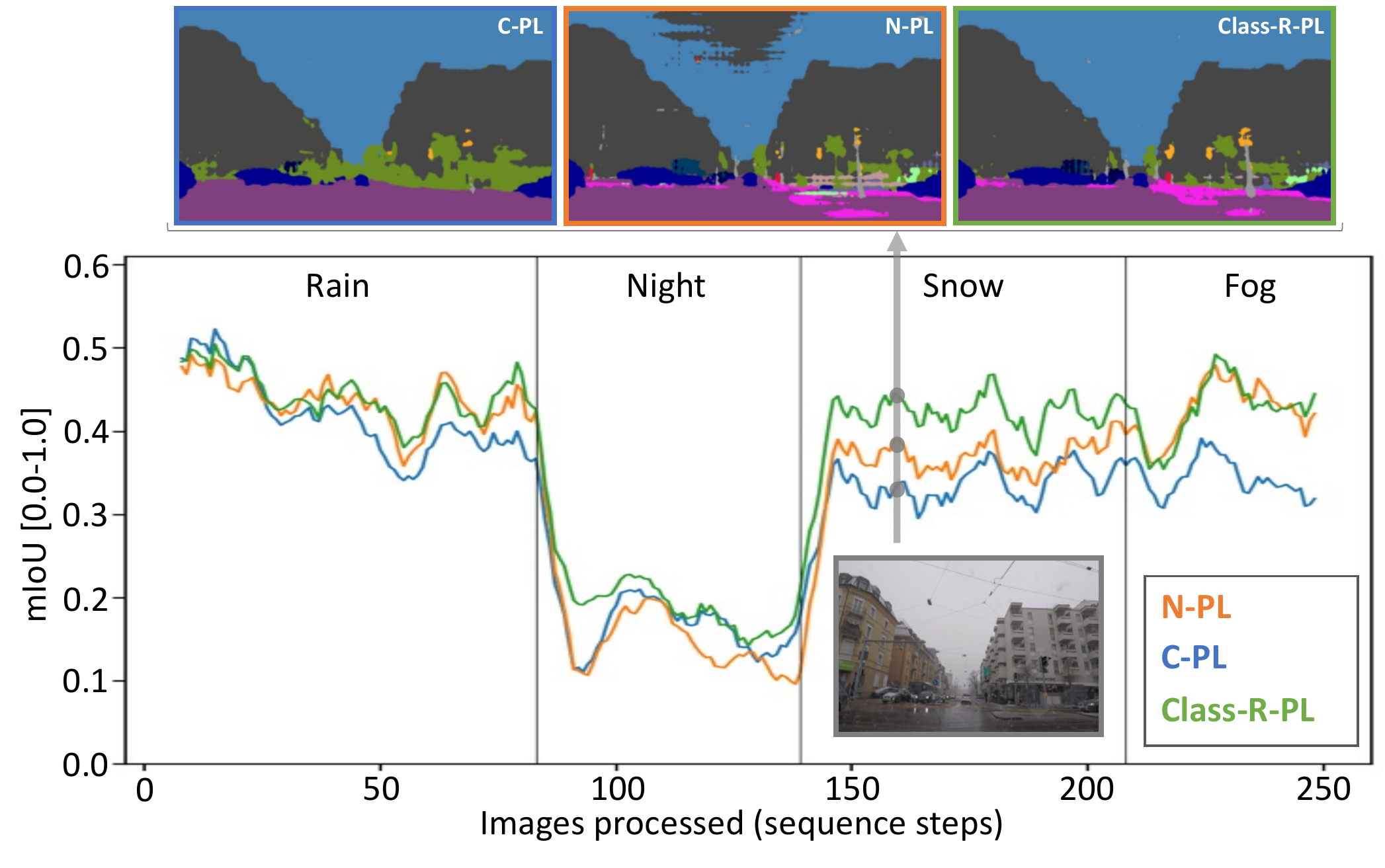}
	\end{center}
	\vspace{-20pt}
	\caption{\footnotesize \textbf{Naive \vs Continual \vs Reset.} Performance evolution of Naive, Vanilla continual, and Class-reset versions of PL (N-PL, C-PL and Class-R-PL, in orange, blue and green, respectively) -- for one ACDC~\cite{Sakaridis21ACDCDataset} sequence. In the plot, we see that learning continuously without precautions results in sub-optimal performance, and that reset allows maintaining a performance close to the naive counterpart in some parts of the sequence, while significantly improving over it in others. Images on top provide a qualitative comparison between predictions of the methods when processing the reported image; C-PL has catastrophically forgotten the classes \textit{sidewalk}, \textit{pedestrian} and \textit{pole} (best viewed in color and zoomed in).
	}
\label{fig:seq_plots}
\vspace{-10pt}
\end{figure}

\myparagraph{Vanilla continual adaptation is not always a safe choice} 
We compare \textit{Naive} (N-PL/N-TENT) adaptation strategies \vs their \textit{Continual} counterparts
(C-PL/C-TENT) in Table~\ref{tab:all_multi}.
These results suggest 
that continual adaptation models may be forgetting important
visual features, resulting in degraded predictions over time (\cf blue and orange curves in Fig.~\ref{fig:seq_plots}). We further report
qualitative evidence of
catastrophic forgetting in Fig.~\ref{fig:seq_plots} (top).

\myparagraph{Using source regularization generally helps}
Introducing a regularization term (C-PL-SR/C-TENT-SR) 
mitigates
the degradation in performance that we
witnessed in some settings (after switching from \textit{Naive} to \textit{Continual} methods).
If we compare the methods with and without SR in Table~\ref{tab:all_multi}, we can observe that adding source regularization always helps the C- methods (\eg, C-PL-SR \vs C-PL). 
We further report in Supplementary
Sec.~\ref{app:results}
qualitative evidences supporting our claim that SR helps mitigating catastrophic forgetting of important information -- while learning continuously. 

\myparagraph{Using reset methods generally helps}
Comparing Class-R- and Oracle-R- methods with their \textit{Naive} (N-) and \textit{Continual} (C-) counterparts
in Table~\ref{tab:all_multi}, we can appreciate how models generally benefit from a reset strategy. As
expected, the oracle outperforms all other methods in every benchmark: while it is not necessarily an upper
bound, it relies on ground truth information, and hence has significant leverage over the other methods. The
gap between Class-R- and Oracle-R- serves as a motivation for future research in reset
methods, \ie different instances of Eq.~\eqref{eq:reset}. Furthermore, note also 
that, while both source regularization and model reset aim at overcoming catastrophic forgetting,
in general the latter provides  stronger improvements. Being complementary methods, they could potentially be combined, yet assessing this is out of scope here.
Finally, Fig.~\ref{fig:seq_plots} 
compares performance over one ACDC sequence
for 
N-PL, C-PL and Class-R-PL (in
orange, blue and green, respectively), showing the catastrophic forgetting avoided with the reset strategy introduced in Section~\ref{sec:contlearning}. 

\section{Concluding remarks}\label{sec:conclusions}

In this paper, we formalize the problem of adapting semantic image segmentation 
models 
% in
to constantly changing environments. 
The resulting task better reflects the challenging conditions that a system typically faces when deployed in the real world: it is exposed to a stream of unlabelled samples whose underlying distribution is constantly shifting.
To study this specific problem, we introduce \textbf{the OASIS benchmark}, which 
% carefully 
separates the validation step from the deploy step,
ensuring that only the best validated model and strategy 
are evaluated on the test set.

Equipped with this benchmark, we borrow approaches from related fields and extend some of them, to serve as diverse baselines for our task, and extensively analyze their behaviors. 
The resulting observations allow us to provide a first answer to the research questions raised in Sec~\ref{sec:protocol_oasis}.
(1)~First, we observe that carefully choosing the right level of domain randomization allows building more robust models that are resilient to the domain shift present in our test sequences, showing that the quality of pre-training should not be neglected. 
(2-3) We compare \textit{naive} 
approaches, which adapt to each 
frame 
restarting from the pre-trained model, with \textit{continual} ones, that adapt continuously. 
Our experiments show that the latter can lead to catastrophic forgetting. 
(4) As a first attempt to mitigate this issue, we propose a reset strategy that carries promising results.  

We believe that addressing the shortcomings of continual, unsupervised
adaptation methods is a promising research path --
towards the deployment of machine learning systems in the real world. 
We hope this benchmark will ease the emergence of new research directions in this sense.

\clearpage

{\small
\bibliographystyle{ieee_fullname}
\bibliography{egbib}
}

\appendix
\clearpage
\newpage
{\Large
\textbf{Supplementary Material}
}

\renewcommand\thefigure{\thesection.\arabic{figure}}    
\setcounter{figure}{0}   
\renewcommand\thetable{\thesection.\arabic{table}}    
\setcounter{table}{0}   

\section{Details of the datasets}\label{app:dataset}

\begin{figure*}
	\begin{center}
    \includegraphics[width=1.0\linewidth]{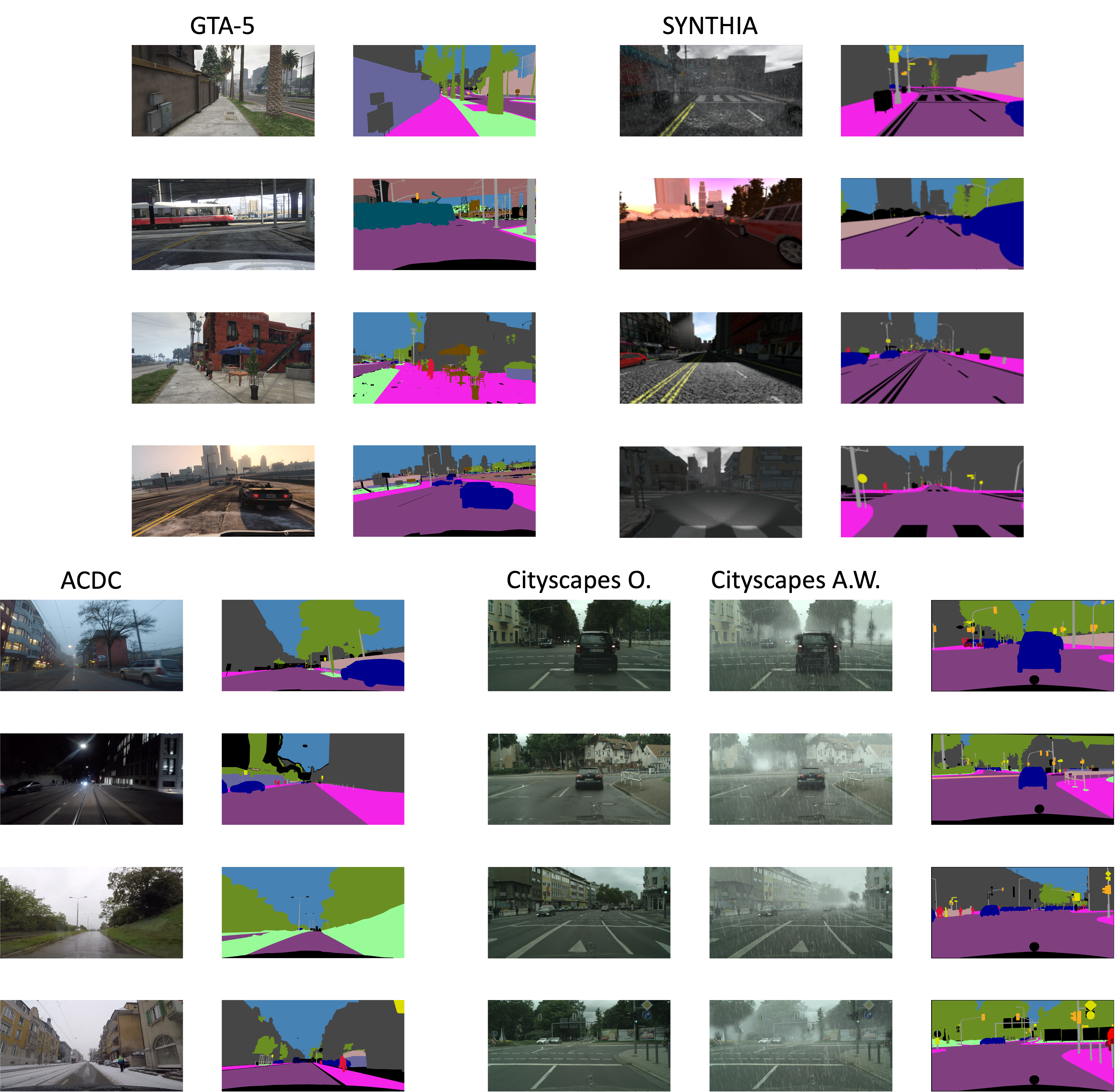}
	\end{center}
	\caption{
	\footnotesize Sample images  of the datasets used for our experiments.}
\label{supp-fig:datasets}
\end{figure*}

We include here additional details related to the datasets we used, namely
GTA-5~\cite{RichterECCV16PlayingForData}, SYNTHIA~\cite{RosCVPR16SYNTHIADataset},
Cityscapes~\cite{CordtsCVPR16CityscapesDataset} and ACDC~\cite{Sakaridis21ACDCDataset}. 
In particular, in Fig.~\ref{supp-fig:datasets} we provide example images  from each dataset  with the corresponding ground-truth segmentation map. Then, in Sec~\ref{sec:classes}
we discuss which categories we considered, as different datasets have 
different annotations available. Finally, we provide details about the sequences we designed for 
the experiments in Sec~\ref{sec:sequences}.

\subsection{Classes}
\label{sec:classes}

In this section, we detail the subset of categories that we considered for our experiments,
%we list the categories we used for our experiments, 
an important point since some classes are available in some datasets but not in others. For example, lane-marking is available in SYNTHIA but not in GTA-5/Cityscapes/ACDC, and vice-versa the terrain class is available in GTA-5/Cityscapes/ACDC but not in SYNTHIA. 
%In the following, we detail the subset of categories that we considered for our experiments. 
We used the 19 classes below from the GTA-5 that are also available in Cityscapes/ACDC: 
%The ones available and used in
%SYNTHIA are highlighted in magenta: 
\textit{\color{magenta}{road}, \color{magenta}{sidewalk}, \color{magenta}{building},
\color{magenta}{fence}, \color{magenta}{pole}, \color{magenta}{light}, \color{magenta}{sign},
\color{magenta}{vegetation}, \color{magenta}{sky}, \color{magenta}{person}, \color{magenta}{car},
\color{magenta}{bicycle}, \color{black}{bus}, \color{black}{train}, \color{black}{motorcycle}, \color{black}{wall}, \color{black}{terrain}, \color{black}{truck}, \color{black}{rider}}. 
% \pdj{I understand you used all 19 classes for GTA-5/Cityscapes/ACDC and only the magenta for SYNTHIA? It is not clear if you only used the magenta for everything. I think I would be a bit more explicit:} 
For GTA-5/Cityscapes/ACDC, we use all of them. For SYNTHIA, we only use the ones highlighted in magenta (the others being unavailable).

\subsection{Sequences}
\label{sec:sequences}

In the following, we report the details of the different sequences used in our experiments. 
This is limited to the SYNTHIA, ACDC, Cityscapes O. and Cityscapes A.W. dataset. In the case of GTA-5, the whole dataset is used for the offline pre-training step; images are randomly sampled from the dataset and the dataset is parsed over several epochs.

\myparagraph{SYNTHIA}
For clarity, this paragraph uses the sequence IDs as reported 
in the dataset's directories: \textit{01} is
\textit{Highway}, \textit{04} is \textit{Old European Town}
and \textit{05} is \textit{New York-like City}. The weather/daylight/seasonal
conditions we used are \textit{Summer},  \textit{Spring}, \textit{Fall},  \textit{Winter}, \textit{Dawn}, \textit{Sunset}, 
\textit{Night}, \textit{Rain}, \textit{Fog}, \textit{Rain-night}, \textit{Winter-night}.
% \pdj{Name also all weather conditions?}
% \gcs{Its maybe a silly suggestion, but it would be nicer to use HW, OTE, NYC instead of 01, 04, ad 05.}
To indicate a sub-sequence shift, we use
arrows ($\rightarrow$). For each sub-sequence, we indicate the
specific environment and the specific weather/daylight condition 
(\eg, \textit{04/Night}). We use $300$ consecutive frames
%samples
per
sub-sequence (to limit the length of the experiments), and build the following sequences -- each one totalling $1.5k$ samples:

\begin{itemize}
    \itemsep0em 
    \small
    \item \textit{05/Night (300 frames)} $\rightarrow$ \textit{01/Dawn (300 frames)} $\rightarrow$ \textit{01/Winter (300 frames)} $\rightarrow$ \textit{05/Winternight (300 frames)} $\rightarrow$ \textit{04/Softrain (300 frames)}
    \item \textit{04/Night (300 frames)} $\rightarrow$ \textit{01/Winter (300 frames)} $\rightarrow$ \textit{04/Softrain (300 frames)} $\rightarrow$ \textit{05/Winternight (300 frames)} $\rightarrow$ \textit{05/Fog (300 frames)}
    \item \textit{01/Winternight (300 frames)} $\rightarrow$ \textit{05/Winternight (300 frames)} $\rightarrow$ \textit{05/Night (300 frames)} $\rightarrow$ \textit{04/Sunset (300 frames)} $\rightarrow$ \textit{04/Winter (300 frames)}
    \item \textit{05/Winternight (300 frames)} $\rightarrow$ \textit{05/Dawn (300 frames)} $\rightarrow$ \textit{05/Night (300 frames)} $\rightarrow$ \textit{05/Sunset (300 frames)} $\rightarrow$ \textit{01/Winter (300 frames)}
    \item \textit{05/Softrain (300 frames)} $\rightarrow$ \textit{01/Night (300 frames)} $\rightarrow$ \textit{04/Fog (300 frames)} $\rightarrow$ \textit{05/Winter (300 frames)} $\rightarrow$ \textit{01/Winternight (300 frames)}
    \item \textit{01/Night (300 frames)} $\rightarrow$ \textit{04/Fog (300 frames)} $\rightarrow$ \textit{01/Fall (300 frames)} $\rightarrow$ \textit{05/Fall (300 frames)} $\rightarrow$ \textit{05/Rain (300 frames)}
    \item \textit{04/Spring (300 frames)} $\rightarrow$ \textit{05/Winter (300 frames)} $\rightarrow$ \textit{04/Night (300 frames)} $\rightarrow$ \textit{01/Dawn (300 frames)} $\rightarrow$ \textit{04/Rainnight (300 frames)}
    \item \textit{01/Winter (300 frames)} $\rightarrow$ \textit{04/Sunset (300 frames)} $\rightarrow$ \textit{04/Spring (300 frames)} $\rightarrow$ \textit{01/Spring (300 frames)} $\rightarrow$ \textit{05/Fog (300 frames)}
    \item \textit{04/Rainnight (300 frames)} $\rightarrow$ \textit{04/Softrain (300 frames)} $\rightarrow$ \textit{05/Winter (300 frames)} $\rightarrow$ \textit{05/Fog (300 frames)} $\rightarrow$ \textit{01/Dawn (300 frames)}
\end{itemize}

\myparagraph{Cityscapes O}
We only use frames from the original Cityscapes dataset
(without weather variations) for which ``fine-grained''
annotation is provided (Cityscapes also provides frames for
which some ``coarse'' annotation is provided). Apart from this,
we do not perform any cut to the sub-sequences, 
% \pdj{what is the length of sub-sequences / sequences here? Also, should we comment here about the fact that sub-sequences are much more similar since we do not have weather changes?}
and build the following sequences:
% \gcs{Should we talk about the fact that we use the segmentation
% dataset where the frames are sampled from the original sequences
% making the continuity between frames less  smooth?}

\begin{itemize}
    \small
    \itemsep0em 
    \item \textit{Aachen (174 frames)} $\rightarrow$ \textit{Hamburg (248 frames)} $\rightarrow$ \textit{Frankfurt (267 frames)} $\rightarrow$ \textit{Munster (174 frames)}
    \item \textit{Jena (119 frames)} $\rightarrow$ \textit{Hamburg (248 frames)} $\rightarrow$ \textit{Zurich (122 frames)} $\rightarrow$ \textit{Hanover (196 frames)}
    \item \textit{Hamburg (248 frames)} $\rightarrow$ \textit{Stuttgart (196 frames)} $\rightarrow$ \textit{Tubingen (144 frames)} $\rightarrow$ \textit{Darmstadt (85 frames)}
    \item \textit{Stuttgart (196 frames)} $\rightarrow$ \textit{Bochum (96 frames)} $\rightarrow$ \textit{Monchengladbach (94 frames)} $\rightarrow$ \textit{Bremen (316 frames)}
    \item \textit{Lindau (59 frames)} $\rightarrow$ \textit{Bochum (96 frames)} $\rightarrow$ \textit{Aachen (174 frames)} $\rightarrow$ \textit{Stuttgart (196 frames)}
    \item \textit{Monchengladbach (94 frames)} $\rightarrow$ \textit{Dusseldorf (221 frames)} $\rightarrow$ \textit{Jena (119 frames)} $\rightarrow$ \textit{Strasbourg (365 frames)}
    \item \textit{Jena (119 frames)} $\rightarrow$ \textit{Strasbourg (365 frames)} $\rightarrow$ \textit{Bochum (96 frames)} $\rightarrow$ \textit{Dusseldorf (221 frames)}
    \item \textit{Strasbourg (365 frames)} $\rightarrow$ \textit{Stuttgart (196 frames)} $\rightarrow$ \textit{Tubingen (144 frames)} $\rightarrow$ \textit{Monchengladbach (94 frames)}
    \item \textit{Krefeld (99 frames)} $\rightarrow$ \textit{Erfurt (109 frames)} $\rightarrow$ \textit{Tubingen (144 frames)} $\rightarrow$ \textit{Strasbourg (365 frames)}
    \item \textit{Monchengladbach (94 frames)} $\rightarrow$ \textit{Lindau (59 frames)} $\rightarrow$ \textit{Aachen (174 frames)} $\rightarrow$ \textit{Jena (119 frames)}
\end{itemize}

\myparagraph{Cityscapes A.W}
We define the following sequences 
%We do not perform any cut to the sub-sequences, and build the following:
by combining sub-sequences (without cut)  from the original Cityscapes~\cite{CordtsCVPR16CityscapesDataset} (Clean), from Cityscapes sequences with artificial \textit{Fog}~\cite{SakaridisECCV2018ModelAdaptSynthRealDataSemDenseFoggySceneUnderstanding} and with artificial \textit{Rain}~\cite{HuCVPR2019DepthAttentionalFeaturesSingleImageRainRemoval}. ``Clean'' indicates that the original sequences are used.

\begin{itemize}
    \itemsep0em 
    \small
    \item \textit{Zurich/Clean (122 frames)} $\rightarrow$ \textit{Darmstadt/Fog (85 frames)} $\rightarrow$ \textit{Dusseldorf/Rain (68 frames)} $\rightarrow$ \textit{Jena/Fog (119 frames)}
    \item \textit{Munster/Rain (30 frames)} $\rightarrow$ \textit{Hamburg/Fog (248 frames)} $\rightarrow$ \textit{Cologne/Clean (154 frames)} $\rightarrow$ \textit{Erfurt/Clean (109 frames)}
    \item \textit{Bremen/Clean (316 frames)} $\rightarrow$ \textit{Stuttgart/Fog (196 frames)} $\rightarrow$ \textit{Aachen/Rain (65 frames)} $\rightarrow$ \textit{Tubingen/Clean (144 frames)}
    \item \textit{Dusseldorf/Rain (68 frames)} $\rightarrow$ \textit{Darmstadt/Clean (85 frames)} $\rightarrow$ \textit{Tubingen/Fog (144 frames)} $\rightarrow$ \textit{Bremen/Rain (53 frames)}
    \item \textit{Bremen/Rain (53 frames)} $\rightarrow$ \textit{Krefeld/Clean (99 frames)} $\rightarrow$ \textit{Lindau/Fog (59 frames)} $\rightarrow$ \textit{Bochum/Clean (96 frames)}
    \item \textit{Cologne/Clean (154 frames)} $\rightarrow$ \textit{Munster/Rain (30 frames)} $\rightarrow$ \textit{Hanover/Fog (196 frames)} $\rightarrow$ \textit{Bremen/Clean (316 frames)}
    \item \textit{Frankfurt/Fog (267 frames)} $\rightarrow$ \textit{Erfurt/Rain (59 frames)} $\rightarrow$ \textit{Zurich/Clean (122 frames)} $\rightarrow$ \textit{Cologne/Clean (154 frames)}
    \item \textit{Hanover/Clean (196 frames)} $\rightarrow$ \textit{Aachen/Fog (174 frames)} $\rightarrow$ \textit{Jena/Fog (119 frames)} $\rightarrow$ \textit{Munster/Rain (30 frames)}
    \item \textit{Bremen/Rain (53 frames)} $\rightarrow$ \textit{Ulm/Fog (95 frames)} $\rightarrow$ \textit{Zurich/Clean (122 frames)} $\rightarrow$ \textit{Darmstadt/Fog (85 frames)}
    \item \textit{Erfurt/Rain (59 frames)} $\rightarrow$ \textit{Ulm/Clean (95 frames)} $\rightarrow$ \textit{Aachen/Rain (65 frames)} $\rightarrow$ \textit{Lindau/Fog (59 frames)}

\end{itemize}

\myparagraph{ACDC}
We do not perform any cut to the sub-sequences, and build the following sequences:% scenarios/sequences/clips?
% \gcs{Do we have the city names for them, would be nicer then  GP010476, GP02047} 
\begin{itemize}
    \itemsep0em 
    \small
    \item \textit{GP010476/Fog (41 frames)} $\rightarrow$ \textit{GP010402/Rain (31 frames)} $\rightarrow$ \textit{GP030176/Snow (22 frames)} $\rightarrow$ \textit{GP010376/Night (56 frames)}
    \item \textit{GP010476/Fog (41 frames)} $\rightarrow$ \textit{GOPR0351/Night (149 frames)} $\rightarrow$ \textit{GOPR0122/Snow (48 frames)} $\rightarrow$ \textit{GP020402/Rain (102 frames)}
    \item \textit{GOPR0402/Rain (83 frames)} $\rightarrow$ \textit{GP010376/Night (56 frames)} $\rightarrow$ \textit{GP010607/Snow (69 frames)} $\rightarrow$ \textit{GOPR0478/Fog (41 frames)}
    \item \textit{GP040176/Snow (86 frames)} $\rightarrow$ \textit{GP020402/Rain (102 frames)} $\rightarrow$ \textit{GP020397/Night (44 frames)} $\rightarrow$ \textit{GP010476/Fog (41 frames)}
    \item \textit{GOPR0122/Snow (48 frames)} $\rightarrow$ \textit{GP020475/Fog (37 frames)} $\rightarrow$ \textit{GOPR0356/Night (50 frames)} $\rightarrow$ \textit{GP010402/Rain (31 frames)}
\end{itemize}

%%%%%%%%%%%%%%%%%%%%%%%%%%%%%%%%%%%%%%%%%%%%
%%%% METHODS %%%%%%%%%%%%%%%%%%%%%%%%%%%%%%%
%%%%%%%%%%%%%%%%%%%%%%%%%%%%%%%%%%%%%%%%%%%%

\section{Details of the methods}\label{app:methods}

\myparagraph{Hyper-parameter selection}
We report details related to the choice of hyper-parameters,
expanding on Section~\ref{sec:exp_details} from the main manuscript. 
We train our models with a
% multi-level
DeepLab-V2~\cite{ChenPAMI17DeeplabSemanticImgSegmentationDeepFullyConnectedCRF}
architecture, implemented in PyTorch~\cite{PaszkeNeurIPS2019PyTorchPaper}. 
We pre-train our models on
GTA-5~\cite{RichterECCV16PlayingForData} for $6$ epochs, using
SGD optimizer with
learning rate $\eta = 2.5\cdot10^{-4}$, momentum $\alpha =
0.9$ and weight decay
$\lambda=5*10^{-4}$. For GPU-memory constraints, we set the batch size to $1$. In the
following, we report the hyper-parameters associated with each method.

\begin{itemize}
    \item \textbf{N-BN:} BN momentum $\alpha=0.1$
    \item \textbf{C-BN:} BN momentum $\alpha=0.1$
    \item \textbf{N-TENT:} learning rate $\eta=1.0$
    \item \textbf{C-TENT:} learning rate $\eta=0.01$
    \item \textbf{C-TENT-SR:} learning rate $\eta=0.01$, source regularizer weight $\gamma=1.0$
    \item \textbf{Class-R-TENT:} learning rate $\eta=0.1$, $K=1$, $\psi=1.0$
    \item \textbf{Oracle-R-TENT:} learning rate $\eta=1.0$
    \item \textbf{N-PL:} learning rate $\eta=10^{-4}$
    \item \textbf{C-PL:} learning rate $\eta=10^{-4}$
    \item \textbf{C-PL-SR:} learning rate $\eta=10^{-4}$, source regularizer weight $\gamma=2.0$
    \item \textbf{Class-R-PL:} learning rate $\eta=10^{-4}$, $K=1$, $\psi=1.0$
    \item \textbf{Oracle-R-PL:} learning rate $\eta=10^{-4}$
\end{itemize}

The hyper-parameters were cross-validated on SYNTHIA sequences,
% \dl{what about the others? Obviously not on the test sets, but still, I would be extra-careful and fill this gap} 
and kept unchanged for ACDC, Cityscapes A.W. and Cityscapes O. sequences.

\myparagraph{Domain randomization (DR)}
To perform domain randomization (DR~\cite{TobinIROS17DomRandTransferDNNSim}), 
we modify the aspect of training samples by applying $K$ different, random image
transformations before feeding each sample to the model. Referring to notation in
Table~\ref{tab:dom_rand} in the main paper, we validate $K=2,3,4$ for 
DR$\uparrow$, DR$\uparrow\uparrow$ and DR$\uparrow\uparrow\uparrow$
%DR+, DR++ and DR+++,
respectively. In the following, we report the different transformations we rely on during
training and we refer the reader to
PIL~\cite{pillow-image-enhance,pillow-image-ops,pil-library} for a more detailed
documentation.

\begin{itemize}
    \itemsep0em 
    \item \textit{Identity:} the image remains unchanged.
    \item \textit{Brightness:} the brightness of the image is 
    perturbed, with intensity in the range  $[0.2; 1.8]$.
    \item \textit{Color:} the color of the image is 
    perturbed, with intensity in the range  $[0.2; 1.8]$.
    \item \textit{Contrast:} the contrast of the image is 
    perturbed, with intensity in the range $[0.2; 1.8]$.
    \item \textit{RGB perturbations:} a random scalar 
    in the range $[0; 120]$ is added to each of the RGB channels.
    \item \textit{RGB-to-gray:} the image is converted to grayscale.
\end{itemize}

%%%%%%%%%%%%%%%%%%%%%%%%%%%%%%%%%%%%%%%%%%%%
%%%% ADDITIONAL RESULTS %%%%%%%%%%%%%%%%%%%%
%%%%%%%%%%%%%%%%%%%%%%%%%%%%%%%%%%%%%%%%%%%%

\section{Additional results}
\label{app:results}

We provide in this section additional results, to extend the main ones included in
the manuscript. To provide a roadmap, %vary; 
in Sec.~\ref{app:results-bn}
we analyze the effect of BN momentum $\eta$ on the N-BN method;
in Sec.~\ref{app:erm-vs-dr-adapt} we show adaptation results obtained when starting from the ERM source model; 
in Sec.~\ref{app:results-adapt-iters} we analyze how the results vary in
%provide results associated with 
the iterative methods N-PL and N-TENT 
when increasing %as 
the number of adaptation iterations;
in Sec.~\ref{app:results-single-cond} we 
provide additional ACDC results, for sequences where only the urban environment
change, but the weather/daylight condition is fixed; 
%provide results associated with increasing magnitube of BN momentum $\eta$, for related to the N-BN method; 
and
finally, in Sec.~\ref{app:results-plots} we provide
more continual learning curves %plots 
such as the one shown in Fig.~\ref{fig:seq_plots} in the main paper. 
% \pdj{Minor comment, it feels weird that we name sec C3 before C1.} %Ric - agree, fixed

% --- BN ---------------------------------------
\subsection{BN adaptation}
\label{app:results-bn}

\begin{figure}[t]
	\begin{center}
      \includegraphics[width=1.0\linewidth]{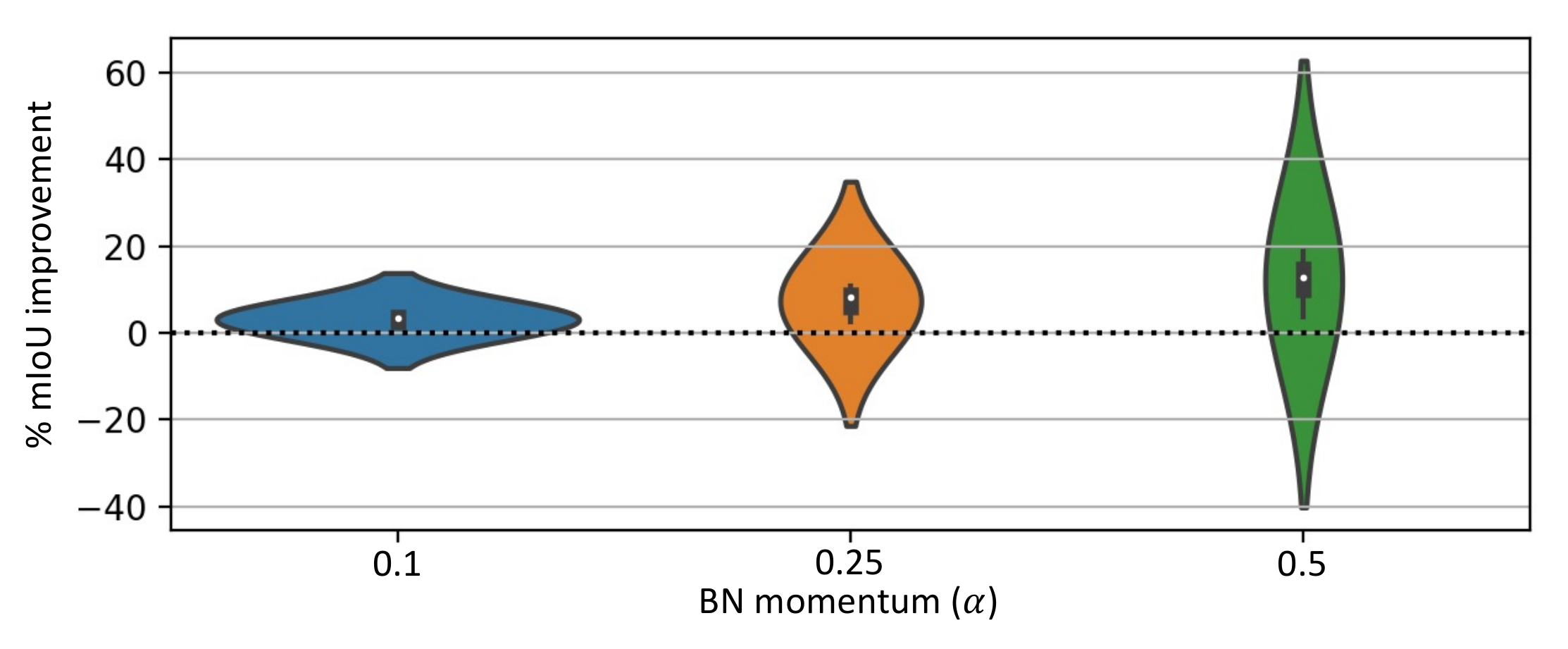}
	\end{center}
	\caption{SYNTHIA results for N-BN -- varying BN momentum ($\alpha$).}
\label{supp-fig:BN-stats-adapt}
\end{figure}

In Figure~\ref{supp-fig:BN-stats-adapt} we showcase the distribution of N-BN results for the
SYNTHIA sequences, when we vary the BN momentum ($\alpha=0.1,0.25,0.5$ in blue, orange and green,
respectively). The general trend is that increasing the momentum $\alpha$ -- that means,
increasing the impact of the target sample's statistics when mixing those with the source
statistics -- leads to higher average results at the price of a significantly larger spread.
These results indicate that the algorithm will perform significantly better on some sequences,
and significantly worse in others. Since this approach is versatile and could be applied in 
tandem with any other method (for example, one could perform continual adaptation with a reset
mechanism, and also adapt BN statistics with increased BN momentum on each sample), we believe
that a more thorough understanding of its behavior on the proposed task %we propose 
represents an interesting research direction.

% --- Adapt from ERM v.s. DR ---------------------
\subsection{Adapting from ERM v.s. adapting from DR}
\label{app:erm-vs-dr-adapt}

In Table \ref{supp-tab:acdc-erm-vs-dr} we compare the adaptation 
results on ACDC between the models started from ERM with the models 
started from DR$\uparrow\uparrow$. We can observe that none of the models started from ERM achieves the performance of the baseline DR$\uparrow\uparrow$ -- apart the Oracle-R, which starting from ERM performs on par with the DR$\uparrow\uparrow$ NA baseline. This confirms the crucial importance of the initial $\mathcal{M}_{\theta_0}$. 
A second observation we can make is that while the numbers are lower for models that start the adaptation from ERM (first column), the overall improvement trend is consistent in general; this emphasizes that the conclusions made concerning the different adaptation strategies hold even if we change the initial $\mathcal{M}_{\theta_0}$.

\begin{table}[t]
\begin{center}
{\scriptsize
\setlength{\tabcolsep}{3.5pt}
\begin{tabular}{@{}lcc@{}}
\multicolumn {3} {c}{\textbf{Comparing adaptation from ERM and DR}} \\
\toprule
& \multicolumn{2}{c}{\textbf{Pre-trained model}} \\
\cmidrule(r){2-3}
\textbf{Method} & \textbf{ERM} & \textbf{DR$\uparrow\uparrow$} \\

\midrule
\textit{\textbf{No adaptation}} & $29.5 $ \tiny {$ \pm  2.5 $ } & $33.6 $ \tiny {$ \pm  2.5 $ }\\
\midrule
\midrule
\textit{\textbf{Style transfer}} & & \\
\midrule
N-ST (NN) &$27.4 $ \tiny {$ \pm  2.2 $ } &$31.9 $ \tiny {$ \pm  2.3 $ }\\
\midrule
N-ST (rand) &$25.5 $ \tiny {$ \pm  1.6 $ } & $31.1 $ \tiny {$ \pm  1.7 $ }\\
\midrule
\textit{\textbf{Naive adapt.}} & & \\
\midrule
N-BN &$30.2 $ \tiny {$ \pm  2.6 $ } &$34.4 $ \tiny {$ \pm  2.5 $ }\\
\midrule
N-PL &$30.4 $ \tiny {$ \pm  2.6 $ } & $34.6 $ \tiny {$ \pm  2.5 $ }\\
\midrule
N-TENT & $31.5 $ \tiny {$ \pm  2.7 $ } &$35.3 $ \tiny {$ \pm  2.7 $ }\\
\midrule
\textit{\textbf{CL adapt.}} & & \\
\midrule
C-BN &$31.7 $ \tiny {$ \pm  2.2 $ } &$35.9 $ \tiny {$ \pm  2.3 $ }\\
\midrule
C-PL & $27.8 $ \tiny {$ \pm  3.4 $ } &$29.7 $ \tiny {$ \pm  3.6 $ }\\
\midrule
C-TENT &$31.2 $ \tiny {$ \pm  2.8 $ } &$34.5 $ \tiny {$ \pm  3.4 $ }\\
\midrule
\textit{\textbf{CL+SR adapt.}} & & \\
\midrule
C-PL-SR &$31.3 $ \tiny {$ \pm  2.8 $ } &$34.5 $ \tiny {$ \pm  2.7 $ }\\
\midrule
C-TENT-SR &$31.1 $ \tiny {$ \pm  2.9 $ } & $35.6 $ \tiny {$ \pm  2.8 $ }\\
\midrule
\textit{\textbf{Adaptive-reset adapt.}} & & \\
\midrule
Class-N-PL &  $32.4 $ \tiny {$ \pm  2.3 $ }&$36.3 $ \tiny {$ \pm  2.3 $ }\\
\midrule
Class-N-TENT &$32.1 $ \tiny {$ \pm  2.3 $ } &$36.0 $ \tiny {$ \pm  2.3 $ }\\
\midrule
\midrule
\textit{\textbf{Oracle-reset adapt.}} & &\\
\midrule
Oracle-N-PL &$33.6 $ \tiny {$ \pm  2.4 $ } &$37.5 $ \tiny {$ \pm  2.3 $ }\\
\midrule
Oracle-N-TENT &$33.6 $ \tiny {$ \pm  2.4 $ } &$37.2 $ \tiny {$ \pm  2.3 $ }\\
\bottomrule
\end{tabular}
}
\end{center}
\caption{Comparison between models adapted starting from ERM or DR$\uparrow\uparrow$ pre-training. Results reported in mIoU.}
\label{supp-tab:acdc-erm-vs-dr}
\end{table}

%\todo{Discuss the table }

% --- More adapt iters --------------------------
\subsection{Adaptation iterations for PL and TENT}
\label{app:results-adapt-iters}
\begin{table*}[t]
\begin{center}
{\scriptsize
\setlength{\tabcolsep}{3.5pt}
\begin{tabular}{@{}lcccccc@{}}
\multicolumn {7} {c}{\textbf{Varying the number of iterations for N-PL and N-TENT}} \\
\toprule
& & & \multicolumn{4}{c}{\textbf{Sequence type}} \\
\cmidrule(r){4-7}
\textbf{Method} & \textbf{Adapt iter.} & \textbf{Learn. rate} & \textbf{SYNTHIA} & \textbf{ACDC} & \textbf{Cityscapes A.W.} & \textbf{Cityscapes O.}\\
\midrule
N-PL & $1$ & $0.0001$ & $+ 3.5 \% $ \tiny {$ \pm  1.0 $ } & $+ 2.9 \% $ \tiny {$ \pm  0.6 $ } & $+ 2.4 \% $ \tiny {$ \pm  1.0 $ } & $+ 1.4 \% $ \tiny {$ \pm  0.2 $ }\\
\midrule
N-PL  & $3$ & $0.0001$ & $+ 7.8 \% $ \tiny {$ \pm  2.7 $ } & $+ 6.5 \% $ \tiny {$ \pm  1.7 $ } & $+ 5.2 \% $ \tiny {$ \pm  2.6 $ } & $+ 2.4 \% $ \tiny {$ \pm  0.4 $ }\\
\midrule
N-PL  & $5$ & $0.0001$ & $+ 9.5 \% $ \tiny {$ \pm  3.7 $ } & $+ 8.4 \% $ \tiny {$ \pm  2.7 $ } & $+ 6.5 \% $ \tiny {$ \pm  3.7 $ } & $+ 2.2 \% $ \tiny {$ \pm  0.6 $ }\\
\midrule
\midrule
N-TENT & $1$ & $1.0$ & $+ 8.5 \% $ \tiny {$ \pm  3.1 $ } & $+ 4.9 \% $ \tiny {$ \pm  2.0 $ } & $+ 3.1 \% $ \tiny {$ \pm  3.6 $ } & $-1.2 \% $ \tiny {$ \pm  0.7 $ } \\
\midrule
N-TENT & $3$ & $1.0$ & $+ 8.4 \% $ \tiny {$ \pm  4.2 $ } & $+ 4.0 \% $ \tiny {$ \pm  3.1 $ } & $+ 3.4 \% $ \tiny {$ \pm  5.3 $ } & $-3.0 \% $ \tiny {$ \pm  1.0 $ }\\
\midrule
N-TENT & $5$ & $1.0$ & $+ 6.8 \% $ \tiny {$ \pm  4.6 $ } & $+ 2.9 \% $ \tiny {$ \pm  3.5 $ } & $+ 2.8 \% $ \tiny {$ \pm  5.9 $ } & $-4.5 \% $ \tiny {$ \pm  1.2 $ }\\
\midrule
\midrule
N-TENT & $1$ & $0.1$ & $+ 4.0 \% $ \tiny {$ \pm  1.1 $ } & $+ 3.1 \% $ \tiny {$ \pm  0.7 $ } & $+ 2.6 \% $ \tiny {$ \pm  1.2 $ } & $+ 1.4 \% $ \tiny {$ \pm  0.2 $ }\\ 
\midrule
N-TENT & $3$ & $0.1$ & $+ 8.9 \% $ \tiny {$ \pm  2.9 $ } & $+ 6.9 \% $ \tiny {$ \pm  1.9 $ } & $+ 5.4 \% $ \tiny {$ \pm  3.0 $ } & $+ 2.1 \% $ \tiny {$ \pm  0.4 $ }\\
\midrule
N-TENT & $5$ & $0.1$ & $+ 10.7 \% $ \tiny {$ \pm  3.8 $ } & $+ 8.5 \% $ \tiny {$ \pm  2.9 $ } & $+ 6.4 \% $ \tiny {$ \pm  4.2 $ } & $+ 1.5 \% $ \tiny {$ \pm  0.6 $ }\\
\bottomrule
\end{tabular}
}
\end{center}
\caption{ Results (relative performance gain in \%) obtained on N-PL and N-Tent when increasing the number of training iterations. }
\label{supp-tab:adapt-iters}
\end{table*}

In Table \ref{supp-tab:adapt-iters} we compare N-PL and N-TENT results 
as we increase the number of iterations.
% where instead of a single training iteration we perform several iterations (3 respectively 5 times),  where
For example, in the case of N-PL we iterate several times between pseudo-labeling and updating the model.
% using the validated learning rate. 
We can observe that, in general, increasing the number of iterations yields better results for N-PL. In the case of N-TENT,
% the best learning rate for single-step adaptation seems to induce overfitting as we increase the number of iterations. Interestingly, 
%where the best learning rate for single iteration was high (1), the model seems overfitting. However,
in the multi-iteration case, results can be significantly improved by reducing the learning rate
(except in the  case of Cityscapes O.). Naturally, this improvement comes with an increased  computational cost, which can be prohibitive according to specific applications -- \eg, autonomous driving. 

%\todo{Discuss the table }

\subsection{Sequences with multiple cities and fixed weather/daylight conditions}
\label{app:results-single-cond}

We report in Table~\ref{supp-tab:acdc-single-cond} results associated with ACDC sequences where
the condition (\textit{Fog}, \textit{Night}, \textit{Rain}, \textit{Snow}) is fixed, and only
the urban environment change. 
% \gcst{The results are averaged over the results obtained for the following sequences: }
For each condition, results are averaged over the following sequences:\\
%We report these sequences below:\\

\textbf{Fog:}

\begin{itemize}
\small
\item \textit{GP020475} $\rightarrow$ \textit{GOPR0478} $\rightarrow$ \textit{GOPR0476} $\rightarrow$ \textit{GP010476}
\item \textit{GOPR0477} $\rightarrow$ \textit{GP020478} $\rightarrow$ \textit{GP010476} $\rightarrow$ \textit{GOPR0475}
\item \textit{GP020475} $\rightarrow$ \textit{GOPR0476} $\rightarrow$ \textit{GOPR0478} $\rightarrow$ \textit{GP010476}
\item \textit{GOPR0477} $\rightarrow$ \textit{GOPR0479} $\rightarrow$ \textit{GOPR0476} $\rightarrow$ \textit{GP020475}
%\item \textit{GP020475 (Fog)} $\rightarrow$ \textit{GOPR0478 (Fog)} $\rightarrow$ \textit{GOPR0476 (Fog)} $\rightarrow$ \textit{GP010476 (Fog)}
%\item \textit{GOPR0477 (Fog)} $\rightarrow$ \textit{GP020478 (Fog)} $\rightarrow$ \textit{GP010476 (Fog)} $\rightarrow$ \textit{GOPR0475 (Fog)}
%\item \textit{GP020475 (Fog)} $\rightarrow$ \textit{GOPR0476 (Fog)} $\rightarrow$ \textit{GOPR0478 (Fog)} $\rightarrow$ \textit{GP010476 (Fog)}
%\item \textit{GOPR0477 (Fog)} $\rightarrow$ \textit{GOPR0479 (Fog)} $\rightarrow$ \textit{GOPR0476 (Fog)} $\rightarrow$ \textit{GP020475 (Fog)}
\end{itemize}

\textbf{Night}
\begin{itemize}
\small
\item \textit{GOPR0351} $\rightarrow$ \textit{GP010376} $\rightarrow$ \textit{GOPR0356} $\rightarrow$ \textit{GP020397}
\item \textit{GP020397} $\rightarrow$ \textit{GOPR0356} $\rightarrow$ \textit{GOPR0376} $\rightarrow$ \textit{GOPR0351}
\item \textit{GP010397} $\rightarrow$ \textit{GP010376} $\rightarrow$ \textit{GOPR0351} $\rightarrow$ \textit{GOPR0356}
\item \textit{GOPR0356} $\rightarrow$ \textit{GOPR0351} $\rightarrow$ \textit{GOPR0376} $\rightarrow$ \textit{GP010376}
%\item \textit{GOPR0351 (Night)} $\rightarrow$ \textit{GP010376 (Night)} $\rightarrow$ \textit{GOPR0356 (Night)} $\rightarrow$ \textit{GP020397 (Night)}
%\item \textit{GP020397 (Night)} $\rightarrow$ \textit{GOPR0356 (Night)} $\rightarrow$ \textit{GOPR0376 (Night)} $\rightarrow$ \textit{GOPR0351 (Night)}
%\item \textit{GP010397 (Night)} $\rightarrow$ \textit{GP010376 (Night)} $\rightarrow$ \textit{GOPR0351 (Night)} $\rightarrow$ \textit{GOPR0356 (Night)}
%\item \textit{GOPR0356 (Night)} $\rightarrow$ \textit{GOPR0351 (Night)} $\rightarrow$ \textit{GOPR0376 (Night)} $\rightarrow$ \textit{GP010376 (Night)}
\end{itemize}

\textbf{Rain}
\begin{itemize}
\small
\item \textit{GOPR0400} $\rightarrow$ \textit{GP020400} $\rightarrow$ 
\textit{GP020402} $\rightarrow$ \textit{GOPR0402}
\item \textit{GP010400} $\rightarrow$ \textit{GP020402} $\rightarrow$ \textit{GOPR0400} $\rightarrow$ \textit{GOPR0402}
\item \textit{GP010400} $\rightarrow$ \textit{GOPR0400} $\rightarrow$ \textit{GP010402} $\rightarrow$ \textit{GP020400}
\item \textit{GOPR0402} $\rightarrow$ \textit{GOPR0400} $\rightarrow$ \textit{GP010400} $\rightarrow$ \textit{GP010402}
%\item \textit{GOPR0400 (Rain)} $\rightarrow$ \textit{GP020400 (Rain)} $\rightarrow$ 
%\textit{GP020402 (Rain)} $\rightarrow$ \textit{GOPR0402 (Rain)}
%\item \textit{GP010400 (Rain)} $\rightarrow$ \textit{GP020402 (Rain)} $\rightarrow$ \textit{GOPR0400 (Rain)} $\rightarrow$ \textit{GOPR0402 (Rain)}
%\item \textit{GP010400 (Rain)} $\rightarrow$ \textit{GOPR0400 (Rain)} $\rightarrow$ \textit{GP010402 (Rain)} $\rightarrow$ \textit{GP020400 (Rain)}
%\item \textit{GOPR0402 (Rain)} $\rightarrow$ \textit{GOPR0400 (Rain)} $\rightarrow$ \textit{GP010400 (Rain)} $\rightarrow$ \textit{GP010402 (Rain)}
\end{itemize}

\textbf{Snow}
\begin{itemize}
\small
\item \textit{GP010607} $\rightarrow$ \textit{GOPR0604} $\rightarrow$ \textit{GOPR0606} $\rightarrow$ \textit{GOPR0122}
\item \textit{GOPR0606} $\rightarrow$ \textit{GP010122} $\rightarrow$ \textit{GP050176} $\rightarrow$ \textit{GOPR0607}
\item \textit{GOPR0607} $\rightarrow$ \textit{GP010122} $\rightarrow$ \textit{GOPR0604} $\rightarrow$ \textit{GP030176}
\item \textit{GP010607} $\rightarrow$ \textit{GP030176} $\rightarrow$ \textit{GOPR0604} $\rightarrow$ \textit{GOPR0606}
%\item \textit{GP010607 (Snow)} $\rightarrow$ \textit{GOPR0604 (Snow)} $\rightarrow$ 
%\textit{GOPR0606 (Snow)} $\rightarrow$ \textit{GOPR0122 (Snow)}
%\item \textit{GOPR0606 (Snow)} $\rightarrow$ \textit{GP010122 (Snow)} $\rightarrow$ \textit{GP050176 (Snow)} $\rightarrow$ \textit{GOPR0607 (Snow)}
%\item \textit{GOPR0607 (Snow)} $\rightarrow$ \textit{GP010122 (Snow)} $\rightarrow$ \textit{GOPR0604 (Snow)} $\rightarrow$ \textit{GP030176 (Snow)}
%\item \textit{GP010607 (Snow)} $\rightarrow$ \textit{GP030176 (Snow)} $\rightarrow$ \textit{GOPR0604 (Snow)} $\rightarrow$ \textit{GOPR0606 (Snow)}
\end{itemize}

\begin{table}[t]
\begin{center}
{\scriptsize
\setlength{\tabcolsep}{3.5pt}
\begin{tabular}{@{}lcccc@{}}
\multicolumn {5} {c}{\textbf{ACDC results (multi environment / single condition)}} \\
\toprule
& \multicolumn{4}{c}{\textbf{Sequence type}} \\
\cmidrule(r){2-5}
& \textbf{Fog} & \textbf{Night} & \textbf{Rain} & \textbf{Snow} \\
 \textbf{No adapt. (NA)}& $41.4 $ \tiny {$ \pm  1.5 $ } & $15.6 $ \tiny {$ \pm  0.9 $ } & $41.9 $ \tiny {$ \pm  0.2 $ } & $38.9 $ \tiny {$ \pm  0.8 $ } \\
\midrule
\textbf{Method} & \multicolumn {4}{c}{\textbf{Improvements}} \\
\midrule
\midrule
\multicolumn{2}{l}{\textit{\textbf{Style transfer}}} & & & \\
\midrule
N-ST (NN) & $-8.8 \% $ \tiny {$ \pm  2.1 $ } & $-3.3 \% $ \tiny {$ \pm  0.5 $ } & $-6.4 \% $ \tiny {$ \pm  0.6 $ } & $-2.8 \% $ \tiny {$ \pm  0.6 $ }\\
\midrule
N-ST (rand) & $-13.9 \% $ \tiny {$ \pm  1.3 $ } & $+ 20.6 \% $ \tiny {$ \pm  1.6 $ } & $-9.9 \% $ \tiny {$ \pm  0.5 $ } & $-10.3 \% $ \tiny {$ \pm  0.5 $ }\\
\midrule
\multicolumn{2}{l}{\textit{\textbf{Naive adaptation}}} & & & \\
\midrule
N-BN & $+ 2.7 \% $ \tiny {$ \pm  0.8 $ } & $+ 4.5 \% $ \tiny {$ \pm  0.3 $ } & $+ 1.6 \% $ \tiny {$ \pm  0.2 $ } & $+ 2.1 \% $ \tiny {$ \pm  0.4 $ } \\
\midrule
N-PL & $+ 3.5 \% $ \tiny {$ \pm  1.1 $ } & $+ 5.1 \% $ \tiny {$ \pm  0.3 $ } & $+ 1.9 \% $ \tiny {$ \pm  0.2 $ } & $+ 2.7 \% $ \tiny {$ \pm  0.4 $ }\\
\midrule
N-TENT & $+ 9.1 \% $ \tiny {$ \pm  3.7 $ } & $+ 5.5 \% $ \tiny {$ \pm  0.7 $ } & $+ 1.2 \% $ \tiny {$ \pm  0.5 $ } & $+ 7.1 \% $ \tiny {$ \pm  1.2 $ }\\
\midrule
\multicolumn{2}{l}{\textit{\textbf{CL adaptation}}} & & & \\
\midrule
C-BN & $+ 8.4 \% $ \tiny {$ \pm  5.1 $ } & $+ 23.7 \% $ \tiny {$ \pm  2.1 $ } & $+ 0.0 \% $ \tiny {$ \pm  0.8 $ } & $+ 6.8 \% $ \tiny {$ \pm  2.1 $ }\\
\midrule
C-PL & $+ 0.8 \% $ \tiny {$ \pm  6.2 $ } & $-41.2 \% $ \tiny {$ \pm  21.1 $ } & $-15.8 \% $ \tiny {$ \pm  2.3 $ } & $-12.2 \% $ \tiny {$ \pm  6.3 $ }\\
\midrule
C-TENT & $+ 8.7 \% $ \tiny {$ \pm  5.7 $ } & $-8.1 \% $ \tiny {$ \pm  2.0 $ } & $-3.0 \% $ \tiny {$ \pm  1.0 $ } & $+ 4.6 \% $ \tiny {$ \pm  2.1 $ }\\
\midrule
\multicolumn{2}{l}{\textit{\textbf{CL+SR adaptation}}} & & & \\
\midrule
C-PL-SR & $+ 8.2 \% $ \tiny {$ \pm  2.6 $ } & $-10.8 \% $ \tiny {$ \pm  1.7 $ } & $-2.0 \% $ \tiny {$ \pm  0.9 $ } & $+ 4.3 \% $ \tiny {$ \pm  2.0 $ }\\
\midrule
C-TENT-SR & $+ 6.1 \% $ \tiny {$ \pm  3.0 $ } & $+ 0.0 \% $ \tiny {$ \pm  3.2 $ } & $+ 0.6 \% $ \tiny {$ \pm  0.1 $ } & $+ 3.5 \% $ \tiny {$ \pm  1.3 $ }\\
\midrule
\multicolumn{2}{l}{\textit{\textbf{Adaptive-reset adaptation}}} & & & \\
\midrule
Class-N-PL & $+ 9.8 \% $ \tiny {$ \pm  5.6 $ } & $+ 25.0 \% $ \tiny {$ \pm  2.2 $ } & $+ 0.2 \% $ \tiny {$ \pm  0.9 $ } & $+ 7.3 \% $ \tiny {$ \pm  2.0 $ }\\
\midrule
Class-N-TENT & $+ 9.4 \% $ \tiny {$ \pm  6.0 $ } & $+ 24.6 \% $ \tiny {$ \pm  2.1 $ } & $-0.4 \% $ \tiny {$ \pm  0.9 $ } & $+ 6.8 \% $ \tiny {$ \pm  2.3 $ }\\
\midrule
\midrule
\multicolumn{2}{l}{\textit{\textbf{Oracle-reset adaptation}}} & & & \\
\midrule
Oracle-N-PL & $+ 13.0 \% $ \tiny {$ \pm  5.9 $ } & $+ 32.2 \% $ \tiny {$ \pm  2.7 $ } & $+ 2.7 \% $ \tiny {$ \pm  0.8 $ } & $+ 10.4 \% $ \tiny {$ \pm  1.9 $ }\\
\midrule
Oracle-N-TENT & $+ 13.6 \% $ \tiny {$ \pm  6.4 $ } & $+ 31.1 \% $ \tiny {$ \pm  2.9 $ } & $+ 1.6 \% $ \tiny {$ \pm  0.8 $ } & $+ 10.1 \% $ \tiny {$ \pm  2.1 $ }\\
\bottomrule
\end{tabular}
}
\end{center}
\caption{ Results (relative performance gain in \%) on ACDC sequences built with multiple   cities  but    fixed weather/daylight conditions.}
\label{supp-tab:acdc-single-cond}
\end{table}

%\todo{Discuss the table + add caption}

\begin{figure*}[!th]
	\begin{center}
      \includegraphics[width=0.9\textwidth]{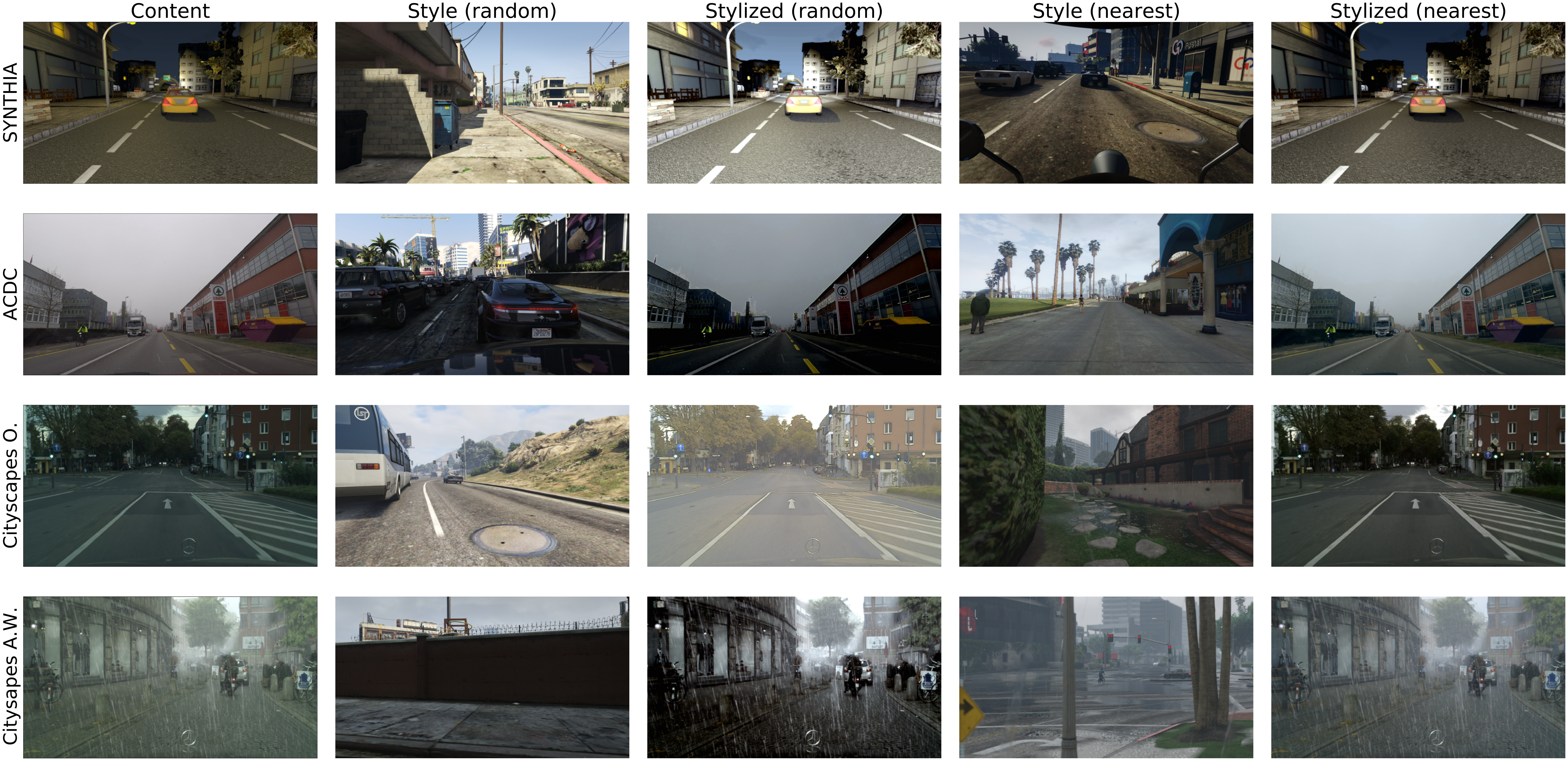}
	\end{center}
	\vspace{-10pt}
	\caption{\small 
	Sample images generated with the style transfer baseline from each dataset. The original \textit{content} image is associated either a random or a nearest neighbour \textit{style} image from the GTA-5 dataset. Then we apply a style transfer method on the content-style pair to obtain a \textit{stylized} image. Note how when choosing style images randomly we can obtain somewhat unrealistic stylized images while choosing them with a simple nearest neighbour search helps mitigate this issue.
	}
	\vspace{-10pt}
\label{supp-fig:style-transfer}
\end{figure*}

\myparagraph{Discussion} When we analyze the results in the Table \ref{supp-tab:acdc-single-cond}, on \textbf{Fog} 
and \textbf{Snow} sequences
we can observe a behavior similar to the one
that we have observed for the ACDC dataset obtained with the multi-environment multi-condition sequences (Table \ref{tab:all_multi}). 
On the other hand, \textbf{Night} and \textbf{Rain} sequences
represent an interesting case study, with some discrepancies.
For example, we can observe significant improvements when adapting with Style Transfer (ST) to Night sequences due to the fact that the ST  can effectively increase the overall brightness and contrast of the image (see Fig.~\ref{supp-fig:style-transfer}), making it easier  for the model to process such samples. 
Furthermore, note that due to a strong domain gap between GTA-5
(containing mainly day images) and images all over these 
sequences  (all night images), the NA baseline yields to a poor performance even with DR$\uparrow\uparrow$.
The continual learning methods C-PL and C-TENT fail %with very low performances 
in this condition, and SR does not carry the same improvements than it did in other setups. The best performing
strategies are C-BN and the reset methods.
% \gcst{This explains probably that in spite of the fact that there is no lighting condition change  (night) observed along the, 
% continuous models without the reset strategies, except the simple C-BN, cannot properly learn to adapt the initial model to 
% this condition, while with the reset strategy we observe a 
% very strong boosting, yielding  the performance  on these hard sequences close to the performance on easier sequences (with a much higher initial DR$\uparrow\uparrow$ performance.}
For the \textbf{Rain} sequences, we observe a similar behavior to the one already observed for Cityscapes O. (Table~\ref{tab:all_multi}). Starting from a strong DR$\uparrow\uparrow$  baseline model,
none of the continual methods,  excluding the Oracle-R approach,  brings significant improvements, and the gain of Oracle-R over the baseline is relatively modest. Surprisingly, there is a significant improvement for the Fog sequence, a condition for which the NA model performs as well as for Rain. Properly understanding different condition-dependent behaviors represents an interesting research question for real-world applications.

% --- Qualitative plots --------------------------
\subsection{Qualitative plots}
\label{app:results-plots}

\begin{figure}[t]
	\begin{center}
      \includegraphics[width=1.0\linewidth]{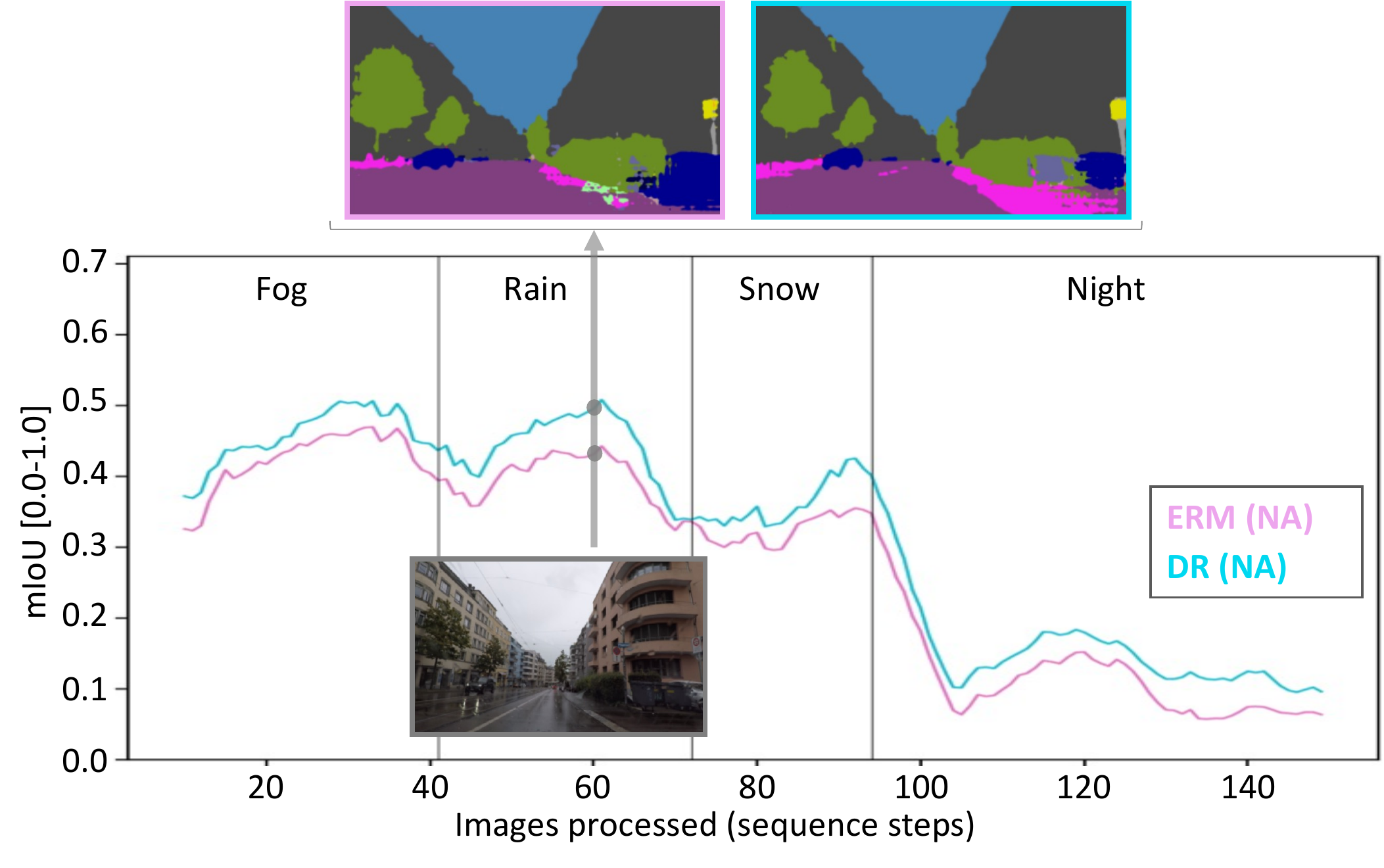}
	\end{center}
	\caption{
	\footnotesize \textbf{ERM \textit{vs} DR} Performance evolution of ERM and DR non-adapted (NA) models,  in pink and light blue, respectively -- for one ACDC~\cite{RosCVPR16SYNTHIADataset} sequence. In the plot, it can be observed that DR brings consistent improvements with respect to ERM. Top: qualitative comparison between predictions of the two methods when processing the reported image; notice the \textit{sidewalk} on the bottom-right corner.
	}
\label{supp-fig:curves-ERM-DR}
\end{figure}
\begin{figure}[t]
	\begin{center}
      \includegraphics[width=1.0\linewidth]{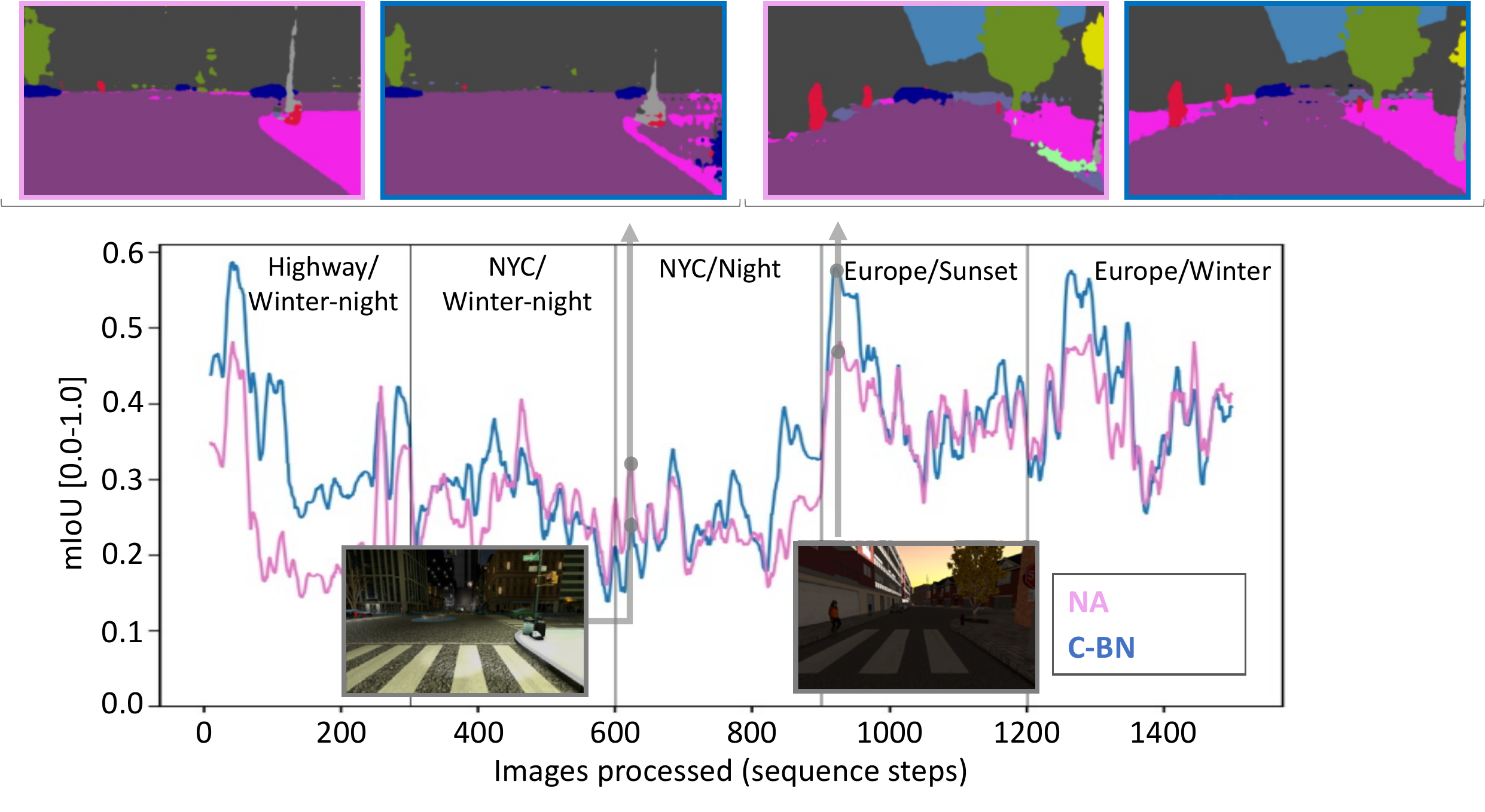}
	\end{center}
	\caption{
    \footnotesize \textbf{C-BN \textit{vs} NA.} Performance evolution of a non-adapted model (NA) and a model whose BN statistics is updated continuously~\cite{ManciniIROS18KittingWildODA}, in pink and blue, respectively -- for one SYNTHIA~\cite{RosCVPR16SYNTHIADataset} sequence. In the plot, it can be observed that C-BN generally brings consistent improvements with respect to NA. Top: qualitative comparison between predictions of the two methods when processing the reported image (on the left, we report failure cases of C-BN w.r.t. NA, from the limited number of frames in which C-BN underperforms.).
	}
\label{supp-fig:curves-BN}
\end{figure}

\begin{figure}[t]
	\begin{center}
    \includegraphics[width=1.0\linewidth]{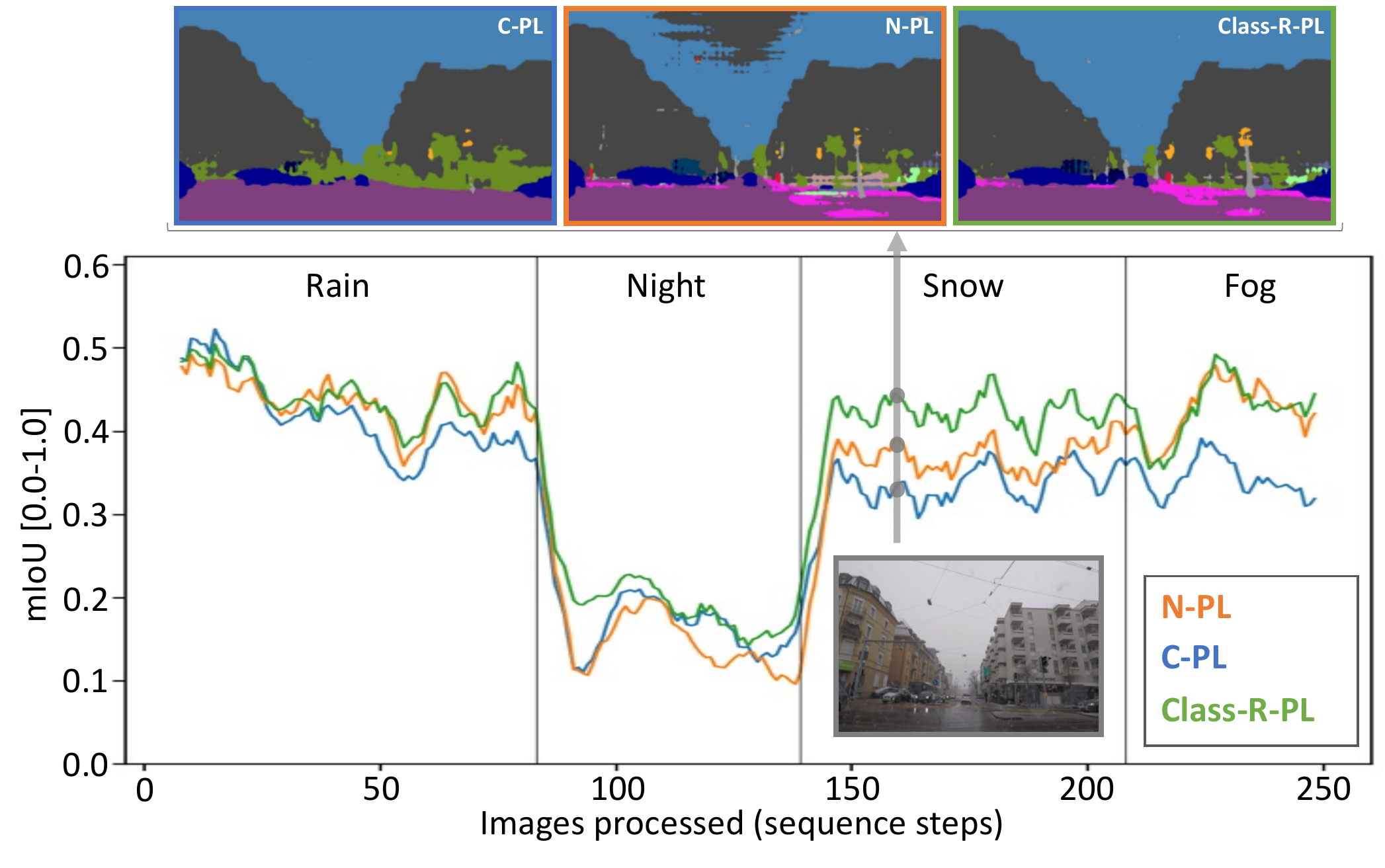}
	\end{center}
	\caption{
	\footnotesize \textbf{Naive \textit{vs} Continual \textit{vs} Reset.} Performance evolution of Naive, ``Vanilla'' continual, and Class-reset versions of PL (N-PL, C-PL and Class-R-PL, is orange, blue and green, respectively) -- for one ACDC~\cite{Sakaridis21ACDCDataset} sequence. In the plot, it can be observed  that learning continuously without precautions results in sub-optimal performance, and that reset can allow maintaining a performance close to the naive counterpart in some parts of the sequence, while significantly improving over it in others. Top: qualitative comparison between predictions of different methods when processing the reported image; C-PL has catastrophically forgotten \textit{sidewalks}, \textit{pedestrians} and \textit{poles} (best in color zooming in).
	}
\label{supp-fig:curves-N-C-R-PL}
\end{figure}

\begin{figure}[t]
	\begin{center}
      \includegraphics[width=1.0\linewidth]{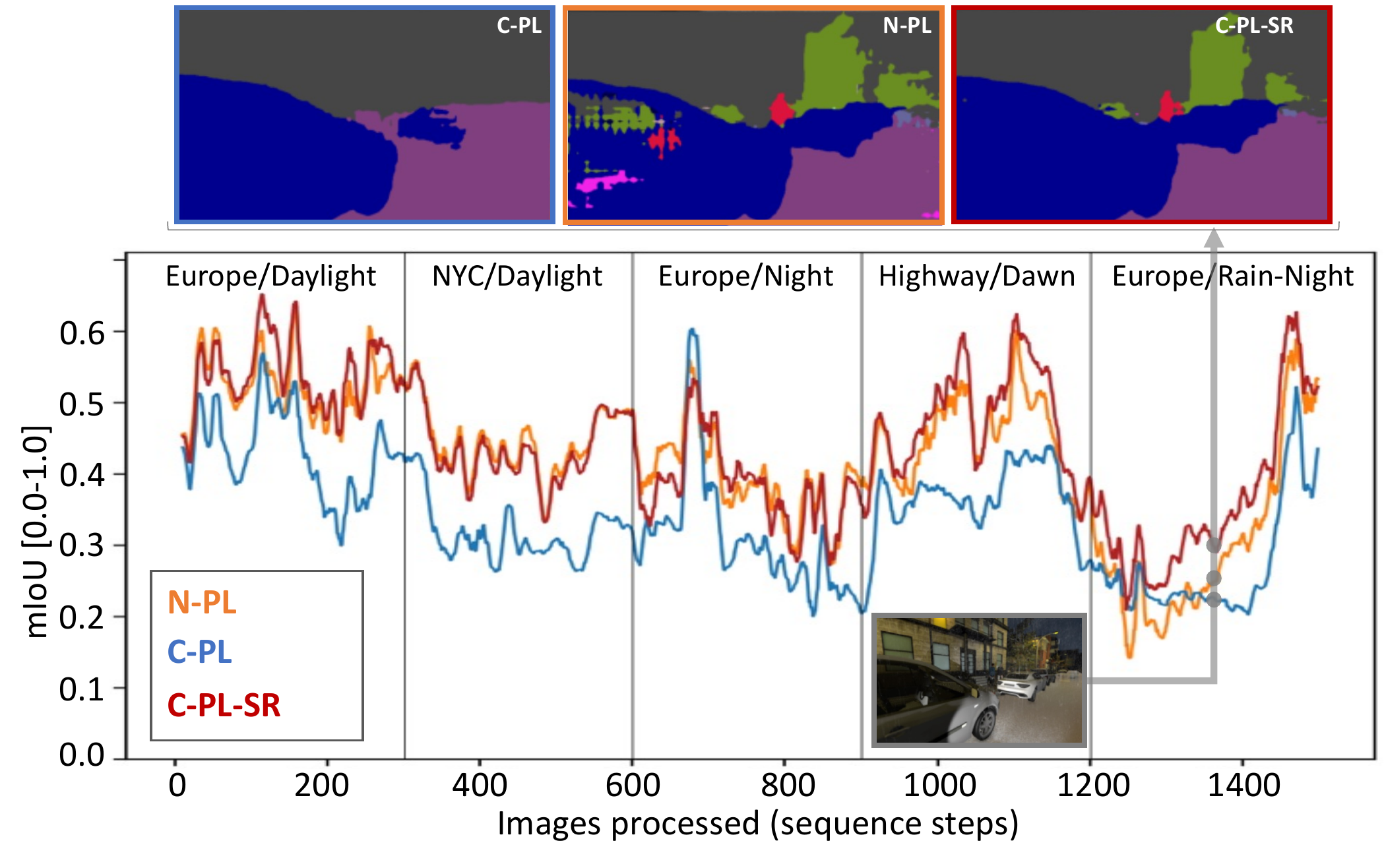}
	\end{center}
	\caption{
	\footnotesize \textbf{Naive \textit{vs} Continual \textit{vs} Source regularized.} Performance evolution of Naive, ``Vanilla'' continual, and Source-regularized versions of PL (N-PL, C-PL and Class-R-PL, is orange, blue and red, respectively) -- for one SYNTHIA~\cite{RosCVPR16SYNTHIADataset} sequence. In the plot, it can be observed  that learning continuously without precautions results in sub-optimal performance, and that source regularization helps mitigating such negative impact, leading to performance often Naive. Top: qualitative comparison between predictions of different methods when processing the reported image; C-PL has catastrophically forgotten \textit{pedestrians} and \textit{vegetation} (best in color zooming in).
	}
\label{supp-fig:curves-N-C-SR-PL}
\end{figure}

We finally report additional qualitative plots, showing the performance evolution
as in Figure~\ref{fig:seq_plots} in the main paper. We report them in Figures~\ref{supp-fig:curves-ERM-DR}---\ref{supp-fig:curves-N-C-SR-PL}, extending
Figure~\ref{fig:seq_plots} from the main manuscript. 
Apart from models reported in Figure~\ref{supp-fig:curves-ERM-DR}, results reported in all other figures assume DR pre-training.
%Apart from models related to the
%Figure~\ref{supp-fig:curves_DR} plot, in all the other figures DR pre-training is assumed.
We summarize the content of each figure below, and report the details in the  different captions.

\begin{itemize}
    \itemsep0em 
    \item \textit{Figure~\ref{supp-fig:curves-ERM-DR}}: We compare models with and without DR~\cite{TobinIROS17DomRandTransferDNNSim} pre-training, in \textit{light blue} and \textit{pink}, respectively.
    
    \item \textit{Figure~\ref{supp-fig:curves-BN}}: We compare models trained with and without BN~\cite{IoffeICML15BatchNormAcceleratingDeepNetsTrain} statistics online adaptation~\cite{ManciniIROS18KittingWildODA}, in \textit{blue} and \textit{pink}, respectively. 
    
    % \item \textit{Figure~\ref{supp-fig:curves-fail}}: We show a failure case of ... \todo{ADD}
    
    \item \textit{Figure~\ref{supp-fig:curves-N-C-R-PL}}:  We compare models trained via N-PL, C-PL and Class-R-PL, in \textit{blue}, \textit{orange} and \textit{green}, respectively.
    
    \item \textit{Figure~\ref{supp-fig:curves-N-C-SR-PL}}:  We compare models trained via N-PL, C-PL and C-PL-SR, in \textit{blue}, \textit{orange} and \textit{red}, respectively.
    
\end{itemize}

\begin{figure*}[!t]
	\begin{center}
      \includegraphics[width=0.8\textwidth]{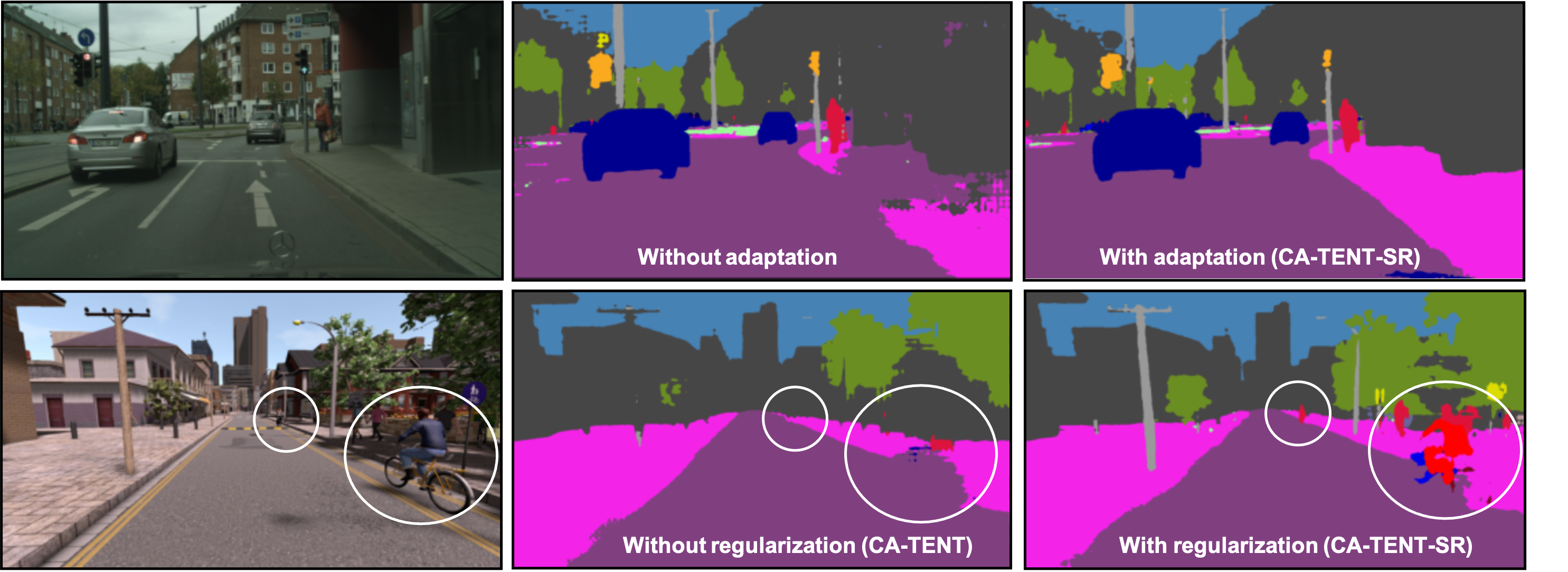}
	\end{center}
	\vspace{-10pt}
	\caption{\small 
	\textit{Top:} We compare the predictions of a non-adapted model \textit{(top-middle)} and a model
	adapted via CA-TENT-SR \textit{(top-right)}. This image shows how the non-adapted model
	struggles in classifying the sidewalk's pixels, while the adapted model improves in this regard.
	\textit{Bottom:} We compare the predictions of models trained via two different versions of the TENT~\cite{WangICLR21TENTFullyTestTimeAdaptEntropyMin} algorithm, when provided with an image \textit{(left)}  CA-TENT (\textit{bottom-middle}) and CA-TENT-SR \textit{(bottom-right)}. This comparison shows how models trained continuously but without regularization can catastrophically forget some classes -- in this case, the pedestrian one.
	}
	\vspace{-10pt}
\label{supp-fig:cat-forg}
\end{figure*}

Figure~\ref{supp-fig:cat-forg} \textit{(top)} provides a qualitative view on improvements led
by adapting the model with C-TENT-SR \textit{(top-right)} with respect to the non-adapted
baseline \textit{(top-middle)}. In this specific example, the improvement in classifying the 
sidewalk's pixels is very significant. Figure~\ref{supp-fig:cat-forg} \textit{(bottom)} provides qualitative evidence of 
% what we intend via ``catastrophic forgetting'' in the main manuscript.
``catastrophic forgetting'', which we discussed about in the main manuscript.
When the model is trained
continuously and without regularization, it can forget classes if it does not encounter
them for a while. In this specific example, the C-TENT model has forgotten the pedestrian
class almost completely \textit{(bottom-middle)}. By regularizing with a term that optimizes the
cross-entropy loss on source samples, the C-TENT-SR model is continuously exposed to the
source classes; %, too; 
hence, it does not forget about them \textit{(bottom-right)}.


%%%%%%%%%%%%%%%%%%%%%%%%%%%%%%%%%%%%%%%%%%%%
%%%% STYLE TRANSFER %%%%%%%%%%%%%%%%%%%%%%%%
%%%%%%%%%%%%%%%%%%%%%%%%%%%%%%%%%%%%%%%%%%%%

\section{Style transfer}\label{app:style-transfer}

One of the most popular approaches for unsupervised domain adaptation is to apply a photorealistic transformation to the source images (usually the training set) so that they resemble the \textit{style} of the target images (usually the test or deployment data) \cite{shrivastava2017learning,WuECCV18DCANDualChannelWiseAlignmentNetworksUDA, ma2018exemplar, cherian2019sem, Richter_2021}. However these methods assume that, even if unlabeled, both the source and target image distributions are known at train time. In this work, we are proposing a benchmark where test images are unknown a priori and we receive them as a continuous stream. Thus, we can not apply these techniques directly. On the other hand, a related line of work has focused on performing single-image style transfer: Given a \textit{content} image and a \textit{style} image, transform the content image to mimic the high-level appearance of the style image.
%(e.g. transform an image taken mid-morning to look as if captured during sunset).
Several works in this area have focused on artistic style transfer, where they provide images with painting styles e.g. \cite{gatys2016image, johnson2016perceptual, huang2017arbitrary, wang2020diversified}. However, when applying these methods between two images they create artifacts that yield unrealistic results. This motivated other works to focus on photorealistic style transfer \cite{luan2017deep, li2018closed, park2019semantic, yoo2019photorealistic, an2020ultrafast}. Although these works were generally motivated from an aesthetics perspective, we note photorealistic style transfer can be used to perform online unsupervised domain adaptation, where for each image received at test time we apply style transfer as a pre-processing step to mimic the appearance of \textit{some} image from the training set. In our work, we use the method described in \cite{yoo2019photorealistic}.

\subsection{Experimental details}
We use the public implementation\footnote{https://github.com/clovaai/WCT2} from \cite{yoo2019photorealistic} with default parameters and stylize at all modules (encoder, decoder and skip connections). We do not use segmentation masks since the content image is unlabeled and using our prediction as a segmentation mask might induce unwanted artifacts. In our experiments, we apply style transfer to every test (or validation) image independently as a pre-processing step. For every content image (test or validation) we choose a corresponding style image (from the training set) to apply the style transfer. We compare two strategies:

\textbf{Random selection:} We select images from the training set uniformly, regardless of the appearance and semantic content of images. This selection method is indeed fast, however, it might pair style and content images which are very different and lead to somewhat artificial image styles.

\textbf{Nearest neighbour selection:} In this case, the idea is to match each content image with the closest style image. We compare images using the cosine similarity between the features extracted after applying a CNN encoder. We note that we can pre-compute the features of the training images offline. Moreover, we use the same encoder used for style transfer as feature extractor for further efficiency. 

Regardless of the selection method, style transfer is a rather expensive procedure which takes of the order of 1-2 seconds per image on a GPU NVIDIA V100, depending on the image size. Therefore, in order to apply it in a real-time scenario, further work should be dedicated into speeding-up the style transfer process. In Figure~\ref{supp-fig:style-transfer} we provide illustrative samples of content, style and stylized images for each dataset and sampling method.

%%%%%%%%%%%%%%%%%%%%%%%%%%%%%%%%%%%%%%%%%%%%
%%%% STREAMLIT APP %%%%%%%%%%%%%%%%%%%%%%%%%
%%%%%%%%%%%%%%%%%%%%%%%%%%%%%%%%%%%%%%%%%%%%

\section{Code and Streamlit web app}
\label{app:code}

Our code to replicate the experiments provided in this work is attached 
to the submission, see the files in \texttt{2419\_code.zip}.

\begin{figure*}[t]
	\begin{center}
      \includegraphics[width=0.8\linewidth]{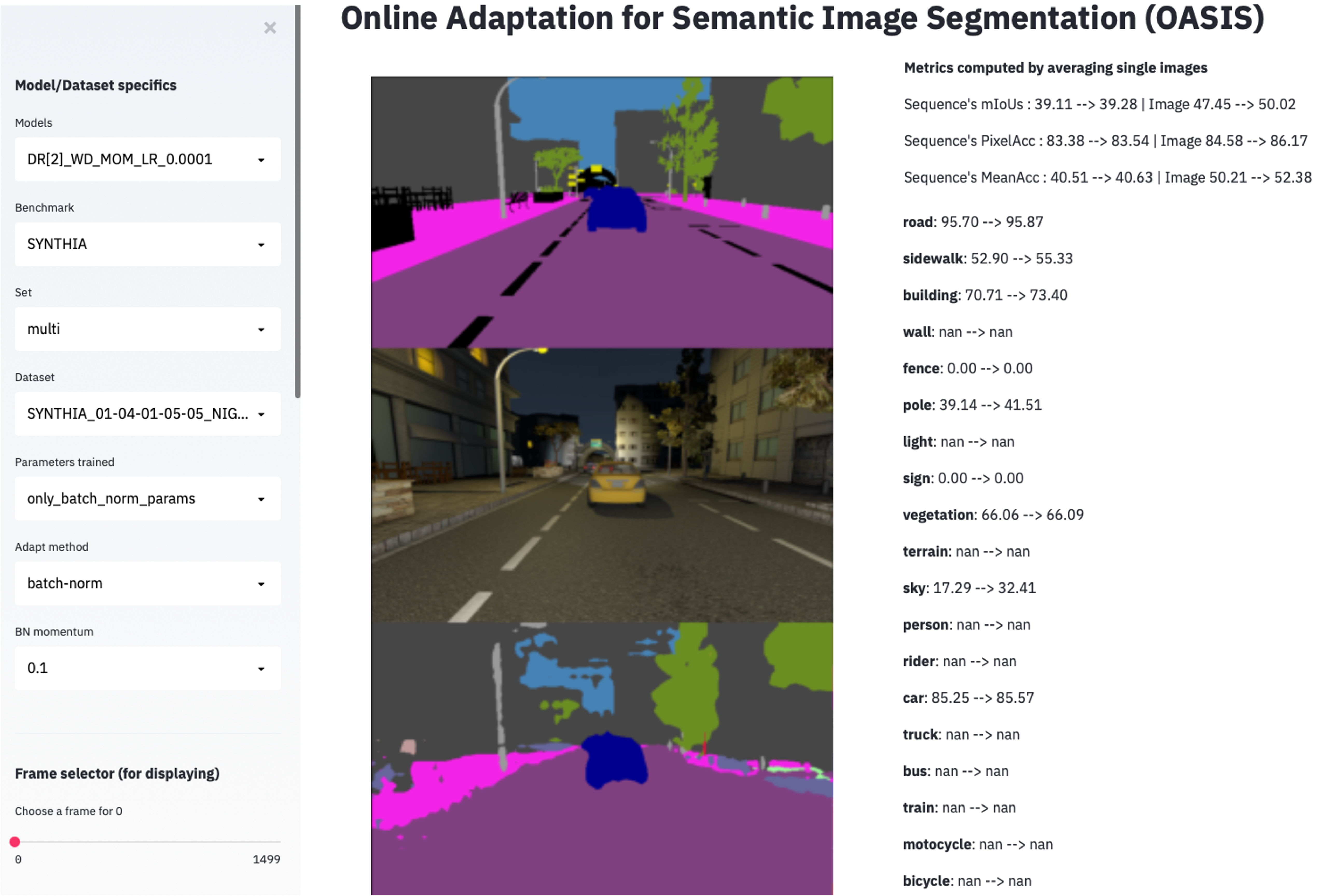}
	\end{center}
	\caption{
	\footnotesize Screenshot of our Streamlit web app.}
\label{supp-fig:streamlit}
\end{figure*}

We report in Fig.~\ref{supp-fig:streamlit} a screenshot of the Web application we will
release to explore results and models (the file used to generate it is
\texttt{streamlit\_app.py}). On the left, one can select the specific model and sequence;
on the right, the results are reported; in the middle, the ground truth, image and
predicted masks are reported -- the selection can be made via the slider on the
bottom-left. We find this app very useful for research on semantic segmentation, where
exploring qualitative results is as important as assessing final performance (\eg,
final mIoU values). The same app can be used to generate the plots shown in Fig.~\ref{supp-fig:curves-ERM-DR}--\ref{supp-fig:curves-N-C-SR-PL}

\section{Limitations}\label{sec:limitations}
We conclude by highlighting the limitations of ideas 
and methods detailed in this work.

For what concerns
the OASIS benchmark we introduced (our core contribution),
we tried, as much as possible, to mimic conditions that
one may face when deploying a machine learning system in the
real world -- in particular, testing on samples/sequences
significantly different from
the ones on which the models have been 
trained and validated.
Yet, it is important to remember
that the real world can expose our models 
to a variety
of conditions significantly broader than the ones a benchmark
can contain. Thus, it is important to avoid having a false sense
of security before deploying a system in the real world; for
example, if we consider an outdoor robot, there may be
combinations of weather/visual conditions
and urban environments, not considered in the benchmark,
that might significantly alter its performance.

For what concerns the methodology, it is important to 
gain familiarity with the failure cases of unsupervised domain
adaptation. As we tried to convey with the experiments reported
in Table~\ref{supp-tab:acdc-erm-vs-dr}, the starting point is
extremely important to foster good adaptation results; if the
pre-trained model is significantly under-performing in some
conditions, domain adaptation can hardly improve on such
performance (in some cases, it can even deteriorate
the performance even more). Finally, 
concerning the implications of continual learning -- and,
more specifically, of continual unsupervised adaptation --
catastrophic forgetting is a severely limiting factor. 
While in Sec.~\ref{sec:contlearning} we have proposed
a family of methods based on a reset mechanism; this is a very 
simple heuristic, far from solving a very broad research problem. 
Still, we hope that raising these limitations will encourage the community 
to consider the problem and devise more advanced solutions to tackle it.

\end{document}